\documentclass[letterpaper, 10 pt, conference]{ieeeconf}  

\IEEEoverridecommandlockouts                              

\usepackage{graphicx}

\usepackage{amsmath} 
\usepackage{amssymb}  
\usepackage[table]{xcolor}
\usepackage{amsmath}
\usepackage{subcaption}
\usepackage{fvextra}
\usepackage{multirow}
\usepackage{multicol}
\usepackage{booktabs}
\usepackage{makecell}
\usepackage{url}
\usepackage{ltxtable}
\usepackage{listings}
\usepackage{comment}
\usepackage{fancyvrb}
\usepackage{caption}
\usepackage{cprotect}
\usepackage{fontawesome5}
\usepackage{bm}

\usepackage{algorithm}
\usepackage{algorithmic}

\DefineVerbatimEnvironment{Verbatim}{Verbatim}{breaklines=true}

\usepackage{inconsolata}
\newif\ifshowcomments
\newif\ifarxiv
\arxivtrue
\showcommentsfalse
\ifshowcomments
    \newcommand{\jz}[1]{{\color{blue}[JZ: #1]}}
   \newcommand{\xt}[1]{{\color{pink}[Xiaoting: #1]}}
   \newcommand{\ry}[1]{{\color{cyan}[RY: #1]}}
   \newcommand{\yanting}[1]{{\color{orange}[YT: #1]}}
    \newcommand{\app}[1]{{\color{red}[#1]}}
\else
    \newcommand{\jz}[1]{{\color{blue}[JZ: #1]}}
   \newcommand{\xt}[1]{{\color{pink}[Xiaoting: #1]}}
    \newcommand{\ry}[1]{{\color{cyan}[RY: #1]}}
   \newcommand{\yanting}[1]{{\color{orange}[YT: #1]}}
   \ifarxiv
    \newcommand{\app}[1]{#1}
    \else
    \newcommand{\app}[1]{}
    \fi
\fi

\usepackage{xcolor}

\usepackage{textcomp}

\author{
Yanting Chen$^{1*\dagger}$, Yi Ren$^{2*}$, Xiaoting Qin$^{2}$, Jue Zhang$^{2\ddagger}$, Kehong Yuan$^{1}$, Lu Han$^{2}$, \\
Qingwei Lin$^{2}$, Dongmei Zhang$^{2}$, Saravan Rajmohan$^{2}$ and Qi Zhang$^{2}$%
\thanks{$^{1}$ Tsinghua University $^{2}$ Microsoft}
\thanks{*Equal Contribution.}
\thanks{\textdagger \textnormal{Work is done during an internship at Microsoft.}}
\thanks{\textdaggerdbl \textnormal{Corresponding author. Contact: jue.zhang@microsoft.com}}
}

\title{
\Large 
\bf \raisebox{-0.3ex}{\includegraphics[height=1em]{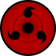}} Sharingan: Extract User Action Sequence from Desktop Recordings}

\begin{document}
\maketitle
\thispagestyle{empty}
\pagestyle{empty}

\begin{abstract}
Video recordings of user activities, particularly desktop recordings, offer a rich source of data for understanding user behaviors and automating processes. However, despite advancements in Vision-Language Models (VLMs) and their increasing use in video analysis, extracting user actions from desktop recordings remains an underexplored area. This paper addresses this gap by proposing two novel VLM-based methods for user action extraction: the Direct Frame-Based Approach (DF), which inputs sampled frames directly into VLMs, and the Differential Frame-Based Approach (DiffF), which incorporates explicit frame differences detected via computer vision techniques. We evaluate these methods using a basic self-curated dataset and an advanced benchmark adapted from prior work. Our results show that the DF approach achieves an accuracy of 70\% to 80\% in identifying user actions, with the extracted action sequences being re-playable though Robotic Process Automation. We find that while VLMs show potential, incorporating explicit UI changes can degrade performance, making the DF approach more reliable. This work represents the first application of VLMs for extracting user action sequences from desktop recordings, contributing new methods, benchmarks, and insights for future research.

\end{abstract}


\vspace{10pt}
\section{Introduction}

Video recordings are increasingly favored for capturing user activities due to their ease of implementation and broad applicability. 
Moreover, video’s universal compatibility across platforms and devices, combined with its ability to capture detailed and context-rich data, ensures minimal information loss and facilitates thorough analysis.

Recent advancements in Vision-Language Models (VLMs)~\cite{GPT-4V, GPT-4o, geminimodels, CLIP, InternVideo, liu2024llavanext, meta2024vlm} have significantly improved the utility of video recordings. These AI-driven models automate the interpretation and extraction of insights from video data, enhancing the identification of user behaviors and patterns. Combined with the increasing prevalence of AI-integrated hardware~\cite{aipin, WearableAI}, these technological innovations are accelerating the adoption of video as an essential tool for documenting and analyzing user activities.

Despite extensive research into understanding user actions from various types of videos~\cite{Video-of-Thought, Wolf}, there remains a lack of focus on desktop recordings. Addressing this gap is crucial, as extracting user actions from desktop videos offers numerous benefits. For instance, it can enhance Robotic Process Automation (RPA) by utilizing demo videos as input, increasing productivity through automation~\cite{smartflow}. Moreover, desktop video analysis facilitates the automatic creation of tutorials and guidelines, while also enabling the extraction of personalized interaction patterns, which can be leveraged elsewhere to create a more personalized user experience.

We propose two VLM-based methods for extracting user action sequences from desktop recordings. In the \textbf{Direct Frame-Based Approach (DF)}, sampled video frames are directly input into VLMs, while the \textbf{Differential Frame-Based Approach (DiffF)} first detects frame changes using computer vision techniques before interpreting them with VLMs. The key difference lies in whether explicit frame differences are incorporated to aid action inference.

We evaluate both methods using two benchmark datasets: one crafted by us, focusing on individual action types, and the other adapted from GUI-World \cite{chen2024guiworld} which better reflects real-world scenarios. Experimental results reveal that current VLMs show great potential in extracting user actions from desktop recordings. For instance, using the DF approach, we achieve an accuracy of $70\% \sim 80\%$ in identifying operation types (e.g., \texttt{click}), and the extracted action sequences are replayable through RPA-like processes. Moreover, comparing the two approaches reveals that VLMs struggle to utilize UI changes derived explicitly, sometimes leading to performance degradation. Thus, we recommend the Direct Frame-Based Approach, relying on VLMs' inherent ability to infer actions, as the current best practice for action extraction.

Our contributions can be summarized as follows:
\begin{itemize}
    \item We introduce two VLM-based methods to address the gap in existing research on extracting user action sequences from desktop recordings. To the best of our knowledge, this is the first attempt to leverage VLMs for this task.
    \item We develop two benchmark datasets to assess the performance of methods on this task. All evaluation source codes and benchmarks will be made publicly available.
    \item We perform a comprehensive evaluation of the proposed methods using the developed benchmarks.
\end{itemize}

\vspace{10pt}
\section{Related Work}
\begin{figure*}[htb!]
    \centering
    \begin{subfigure}[b]{0.45\textwidth}
        \centering
        \includegraphics[width=0.8\textwidth]{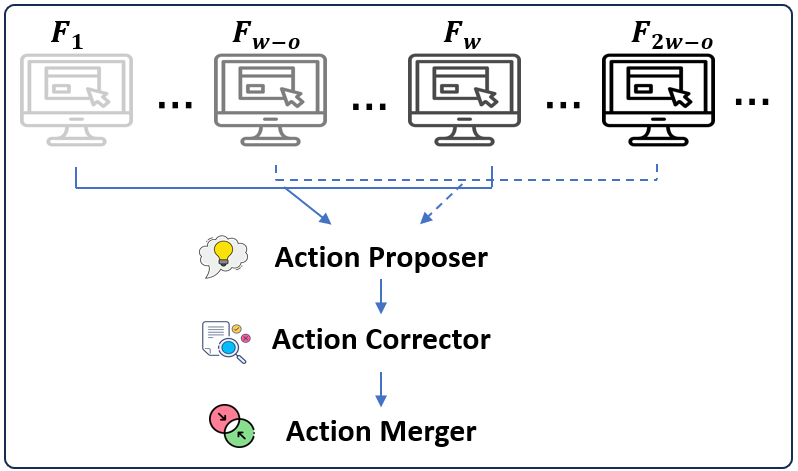} 
        \caption{Direct Frame-Based Approach}
        \label{fig:DF_arch}
    \end{subfigure}
    \hspace{-0.1\textwidth} 
    \begin{subfigure}[b]{0.45\textwidth}
        \centering
        \includegraphics[width=0.8\textwidth]{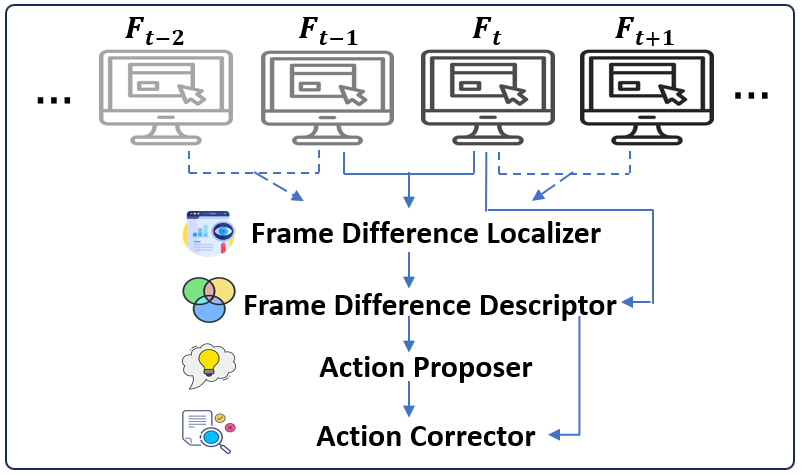} 
        \caption{Differential Frame-Based Approach}
        \label{fig:DiffF_arch}
    \end{subfigure}    
    \caption{Architectures of Direct Frame-Based Approach (left) and Differential Frame-Based Approach (right).}
    \label{fig:DF_DiffF_arch}
\end{figure*}

\noindent \textbf{VLMs in Robotics and Automation}
VLMs have been increasingly applied across various robotics and automation tasks. They have proven valuable for robot navigation \cite{liu2024VLnavigation, Zhang24Navigation, Yokoyama24Navigation}, robot action recognition \cite{Wei24Action, Deng24Recognition}, and task planning \cite{Shirai2024VLrobot}, contributing significantly to enhancing robotic visual perception capabilities \cite{Kapelyukh2024VL3D, Long2024robollm}. These works primarily focus on physical environments where VLMs help robots interpret and interact with the world around them, and specific applications like autonomous driving \cite{Wang23drive} exemplify their utility in high-stakes scenarios. Our work focuses on robotic process automation in desktop environments. Recent studies, such as GUI Agent-related works \cite{OmniParser, ufo, cradle, VisualWebBench}, demonstrate the potential of VLMs in automating user interface tasks by processing natural language inputs. These approaches largely ignore the vast potential of video as a data modality. Our study is orthogonal but distinct—by focusing on extracting user action sequences directly from desktop video recordings, we enable new downstream tasks such as task automation, personalized software tutorials, and workflow optimization in desktop RPA from demo videos.



\noindent \textbf{VLM-based Video Understanding}
The integration of large language models (LLMs) and VLMs into video understanding frameworks has seen rapid advancements. Existing approaches typically address video understanding by querying sampled frames and generating captions to retrieve visual information or answer questions~\cite{VideoTree, VideoAgent1, videoagent, VideoChat}. However, these works often remain focused on image-based tasks, failing to fully exploit the temporal dynamics of videos or tackle action-related tasks in complex settings. Our focus diverges from these existing works by emphasizing action recognition in dynamic video sequences, particularly within desktop environments where user activities are rich in temporal complexity. Recent efforts like Video-of-Thought \cite{Video-of-Thought} and Wolf \cite{Wolf} have begun to explore spatial-temporal understanding, but their experiments center on human action recognition in daily life or physical activities.
By introducing a dataset tailored for desktop environments, our framework and rigorous evaluation not only bridges the gap in extracting actionable insights from desktop recordings but also serves as a stepping stone for future research in automating user-centric desktop tasks with video as a primary input modality.

\vspace{10pt}
\section{methodology}


This work focuses on extracting sequences of user actions from desktop video recordings by leveraging VLMs that analyze videos at the frame level.\footnote{The inclusion of audio modality is beyond the scope of this work.} To achieve this, we uniformly sample $n$ frames $\mathcal{F} = \{F_t\}$ from the video $\mathcal{V}$, where the frame $F_t$ is selected at time $t$. The extracted action sequence, represented as $\mathcal{A} = \{A_i\}$, consists of individual actions $A_i$ described textually at step $i$. Each action $A_i$ is characterized by a tuple $(O_i, D_i, C_i)$, where $O_i$ represents the operation type (e.g., \texttt{click}), $D_i$ details the operation specifics (e.g., the UI element interacted with), and $C_i$ provides the context (e.g., the application in use). This study considers five operation types: \texttt{click}, \texttt{select}, \texttt{scroll}, \texttt{drag}, and \texttt{type}. An illustrative example of an action is \emph{(``click'', ``Styles dropdown menu'', ``Microsoft Word'')}. 


We explore two methodologies for extracting actions from sampled video frames. The first, the \textbf{Direct Frame-Based Approach (DF)}, directly inputs sampled frames into VLMs to generate an action sequence. This approach necessitates that VLMs adhere to specific instructions embedded in the prompts, which are crafted using prompt tuning techniques such as in-context learning and chain-of-thought reasoning.
In contrast, the \textbf{Differential Frame-Based Approach (DiffF)} first utilizes computer vision techniques to explicitly identify regions of change between consecutive frames. VLMs are then used to interpret these changes and generate the action sequence. The primary distinction between these methods lies in whether the frame differences are explicitly extracted, enabling an investigation into whether this extraction enhances action inference by VLMs. The following sections provide a detailed description of each approach, with their architectures depicted in Figure~\ref{fig:DF_DiffF_arch}.



\subsection{Method I: Direct Frame-Based Approach}

As shown in Figure~\ref{fig:DF_arch}, DF consists of three modules: Action Proposer, Action Corrector and Action Merger. Since current VLMs often only handle a limited number of images in one call, we use a sliding window to process $w$ frames at a time, allowing $o$ overlapping frames. The Action Merger combines results from all windows. If frames can fit within the VLM's context, the sliding window and Action Merger are unnecessary. Each module is described in detail below.

\noindent\textbf{Action Proposer.} This module proposes a candidate action sequence for sampled frames within each time window. The prompt\footnote{Due to space limit, this and following prompts are not included in current draft. We intend to open source them in supplementary materials.} used for this task\app{, detailed in Table \ref{prompt:action-proposer-df-1} in the Appendix,} directs the VLM to focus on changes potentially related to user actions. For instance, when inferring a \texttt{click}, the VLM utilizes several features: i) mouse shape change, such as an arrow-like pointer shifting to a hand when hovering over a clickable element; ii) changes in the UI element's state, like background color adjustments when the mouse hovers; and iii) new events triggered, such as opening a new window or expanding a menu. By synthesizing these visual cues across frames, the VLM determines the user’s action type $O$ as \texttt{click}, identifies the clicked UI element as the action detail $D$, and specifies the application name as context $C$. Additionally, the VLM generates intermediate outputs, including mouse position, pre- and post-action element states, and its reasoning process. These outputs support the Action Corrector in performing error corrections.

\noindent\textbf{Action Corrector.} This module corrects several types of errors presented in Action Proposer's output, including: i) redundant actions, such as an unnecessary \texttt{click} action accompanying a \texttt{drag} operation; ii) invalid actions, where the operation type, details, or context are inconsistent with other supporting information; and iii) missing information, such as incomplete operation details or context, which might be inferred from supplementary data. \app{The prompt for this module is provided in Table \ref{prompt:corrector-df-1} in the Appendix.}

\noindent\textbf{Action Merger.} This module merges the generated action sequences across all sliding time windows, addressing issues related to fragmented operations from video slicing and redundant actions caused by overlapping windows. For example, consecutive \texttt{select} actions on text are consolidated into a single \texttt{select} action by merging the operation details from each individual action. \app{The prompt for this module is detailed in Table \ref{prompt:merger-df-1} in the Appendix.}

\subsection{Method II: Differential Frame-Based Approach}

The DiffF approach, illustrated in Figure~\ref{fig:DiffF_arch}, consists of four components: Frame Difference Localizer, Frame Difference Descriptor, Action Proposer and Action Corrector. In contrast to the DF approach, DiffF introduces the Frame Difference Localizer and Descriptor to explicitly capture frame differences. The Action Proposer and Corrector modules, while playing similar roles to those in DF, are adapted to the DiffF framework. Note that DiffF bypasses previous input image limits for VLMs as frame difference generation requires only two consecutive frames. Additionally, since action proposing and correction are handled in text and VLMs typically support extensive textual context, Action Merger is generally unnecessary in DiffF, except for exceptionally long videos, which are not addressed in this work.

\noindent\textbf{Frame Difference Localizer.} This module identifies and outputs screen regions that have changed between two consecutive frames. \app{An example of the detected regions with annotated changes is illustrated in Fig. \ref{fig:comparator-example} in the Appendix. }The localization process involves the following steps with the tools~\cite{opencv_library, scikit-image}:
\begin{itemize}
\item Normalize the RGB pixel values of input frames to [0, 1]. 
\item Apply a Gaussian blur with a $(5, 5)$ kernel size and a standard deviation of 2 to reduce false-positive UI changes caused by high-frequency noise, commonly introduced by lossy video compression. This configuration is found to be effective at minimizing false positives while preserving essential, albeit minor, UI changes.
\item Calculate the L2 norm of the difference in pixel values between the two frames and threshold the resulting difference with a tuned value of $0.15$ to create a binary mask. 
\item Remove objects smaller than 10 pixels from the binary mask (as changes smaller than this threshold are generally imperceptible to the human eye) and identify connected components with their corresponding bounding boxes. 
\item Expand each bounding box by 100 pixels on all sides and merge any overlapping boxes to provide more visual context while reducing the number of regions to compare.
\end{itemize}

\noindent \textbf{Frame Difference Descriptor.} This module utilizes a VLM to  generate detailed textual descriptions of UI changes based on current frame and detected changed regions as compared to previous frame in the above step. \app{The prompt for this module is provided in Table \ref{prompt:frame-difference-descriptor-1} in the Appendix, along with a sample output corresponding to the changes illustrated in Figure~\ref{fig:comparator-example}. }The output includes overall frame context and specific details about the changed UI elements. However, since  Descriptor is only provided with localized information (i.e., current frame and their frame differences), it does not generate action sequences directly. 


\noindent \textbf{Action Proposer.} After aggregating textual descriptions of UI changes across all frames, this module prompts the VLM to propose candidate actions, forming an action sequence for the entire video. The prompt used in this process is analogous to that in the DF approach and differs mainly in the supporting information part, which will be utilized in the next Action Corrector module. \app{A detailed description of the prompt is provided in Table~\ref{prompt:action-proposer-1} in the Appendix.}

\noindent \textbf{Action Corrector.} Given that DiffF is prone to generate false positive actions due to extraneous information in the textual descriptions of UI changes, the Action Corrector incorporates an additional rule-based component not present in DF. While the VLM-based corrector in DiffF employs a similar prompt to the Action Corrector in DF, the rule-based component specifically targets the elimination of \texttt{scroll} actions without corresponding UI movement and \texttt{click} actions where the cursor is absent from the evidence.

\vspace{10pt}
\section{Evaluation}

In this section, we first introduce two benchmark datasets \textsc{ActOne} and \textsc{ActReal}, followed by a description of the evaluation methods and metrics used in the experiments.

\subsection{Benchmark Datasets}

\begin{table}[t]
  \centering
  \scriptsize
  \begin{tabular}{c | c | c | c | c | c}
  \hline
  \hline
  Dataset & \makecell[c]{Case\\Domain} &\makecell[c]{Total\\Videos} & \makecell[c]{Frame\\Count} & \makecell[c]{Action\\Count} & \makecell[c]{Unique Action\\Type Count} \\
  \hline
  \multirow{6}{*}{\makecell[c]{\textsc{ActOne}}} & \texttt{click} & 14 & 276 & 1.7 & 1.1\\
  & \texttt{select} & 11 & 242 & 1.0 & 1.0 \\
  & \texttt{scroll} & 6 & 292 & 1.2 & 1.2 \\
  & \texttt{drag} & 5 & 285 & 1.4 & 1.4 \\
  & \texttt{type} & 4 & 255 & 1.5 & 1.3 \\
  \cline{2-6}
  & \textbf{All} & 40 & 277 & 1.3 & 1.2 \\
  \hline
  \multirow{4}{*}{\makecell[c]{\textsc{ActReal}}} & Software & 23 & 684 & 7.5 & 3.0 \\
  & Website & 15 & 985 & 8.3 & 3.1\\
  & Multi & 3 & 836 & 6.0 & 3.3 \\
  \cline{2-6}
  & \textbf{All} & 41 & 805 & 7.7 & 3.1   \\
  \hline
  \hline
  \end{tabular}
  \caption{Statistics of benchmark datasets \textsc{ActOne} and \textsc{ActReal}. The numbers in the last three columns are the average values computed within each respective case domain.}
  \label{tb:dataset_meta}
\end{table}

\noindent \textbf{\textsc{ActOne.}} Using OBS Studio~\cite{obs} for screen recording and manual annotation of action sequences, we built the \textsc{ActOne} dataset, specifically designed to evaluate VLMs' fundamental abilities in action sequence extraction tasks. This dataset includes five operation types of (\texttt{click}, \texttt{select}, \texttt{scroll}, \texttt{drag}, and \texttt{type}), covering common usage scenarios in desktops, web browsers and popular applications. 
Since the goal is to assess fundamental capabilities, the videos do not contain overly complex actions; most of them only include a single action type. Additionally, as current VLMs often allows no more than ten images per call, to ensure a fair comparison across different models, all videos in this dataset are limited to a maximum duration of 10 seconds. Details on video and action statistics is provided in Table~\ref{tb:dataset_meta}.

\noindent \textbf{\textsc{ActReal.}} 
To evaluate real-world user activities, we construct the \textsc{ActReal} dataset adapted from GUI-World \cite{chen2024guiworld}, a collection of videos sourced from YouTube. We select videos from the ``Software'', ``Website'', and ``Multi'' categories to align with our focus on desktop recordings. Videos containing hover actions or non-input keyboard actions are filtered out, and we limit the dataset to videos with 6-10 actions and at least 3 unique action types to ensure diversity and complexity. After filtering, 36 videos remain in the ``Website'' category, 12 in ``Multi'', and 186 in ``Software''. Upon manual review, many videos exhibit quality issues like cropped displays, low resolution, obstructions from watermarking, or incomplete/wrong annotations. We discard these videos and manually re-annotate the action sequences for the remaining ones. The final dataset consists of 41 videos. As shown in Table~\ref{tb:dataset_meta}, compared to \textsc{ActOne}, the average number of actions per video increases approximately five-fold, and the number of unique actions per video is tripled, making this dataset significantly more complex and diverse.

\subsection{Evaluation Methods and Metrics}

We primarily evaluate the predicted action sequences by comparing them with the ground-truth sequences in the benchmark datasets. Although the comparison is performed in semantic space rather than through exact word matching, errors still arise due to the variability in how the same UI object can be described in different ways. To further validate the proposed semantic comparison metrics, we introduce another functional metric derived by replaying the predicted action sequences using a VLM-based UI automation tool in the same environment as the benchmark dataset. If the final outcome matches the ground-truth video, it confirms the accuracy of the action sequence. This replay approach mirrors the Robotic Process Automation (RPA) process, a key potential application of this work. Below, we provide a detailed description of both evaluation methods.

\noindent\textbf{Semantic Comparison.} The comparison is performed at the individual video level by analyzing two sequences of text strings: the predicted action sequence, $\mathcal{A}^p = \{ (O_i^p, D_i^p, C_i^p) \mid i \in \mathcal{L}^p \}$, and the ground-truth action sequence, $\mathcal{A}^g = \{ (O_j^g, D_j^g, C_j^g) \mid j \in \mathcal{L}^g \}$. These sequences may have differing lengths, as denoted by $|\mathcal{L}^p|$ and $|\mathcal{L}^g|$. 

Our comparison process begins by computing three similarity matrices, $S_{ij}^O$, $S_{ij}^D$, and $S_{ij}^C$, for each action element, where $i \in \mathcal{L}^p$ and $j \in \mathcal{L}^g$. Since the operation type $O$ is discrete and limited to five categories, $S_{ij}^O$ is computed via exact matching, producing a binary (0-1) matrix. For $S_{ij}^D$ and $S_{ij}^C$, semantic matching is performed by: i) generating BERT embeddings~\cite{bert} for the operation detail $D$ and context $C$ elements from both the prediction and ground-truth sets; ii) computing pairwise cosine similarity between the embeddings; iii) applying a manually tuned threshold of 0.7 to convert similarity scores into a binary matrix. Once the binary similarity matrices for all three action components are obtained, the overall similarity matrix $S_{ij}$ is calculated through element-wise multiplication.

We next perform a matching between $\mathcal{A}^p$ and $\mathcal{A}^g$ and count the matched pairs for metrics computation. The matching algorithm involves three steps: i) iterating over ground-truth actions in chronological order; ii) for each ground-truth action, identifying the first unmatched predicted action that aligns with it (indicated by a $1$ in $S_{ij}$), and treating them as a match; iii) counting the number of matched pairs $m$. 


\begin{algorithm}
\caption{Matching algorithm}
\begin{algorithmic}
\STATE \textbf{Input:} $S_{ij}$
\STATE \textbf{Output:} number of matched pairs $m$
\STATE
\STATE $M \leftarrow [\text{False for } i = 1\ldots |\mathcal{L}^p|]$: $M[i]$ represents whether $\mathcal{A}^p_{i}$ has been matched previously
\STATE $m \leftarrow 0$
\FOR{$j \leftarrow 1$ to $| \mathcal{L}^g|$}
    \FOR{$i \leftarrow 1$ to $|\mathcal{L}^p|$}
    \IF{$M[i] = $ False and $S_{ij} = 1$}
        \STATE $m \leftarrow m + 1$
        \STATE $M[i] \leftarrow $ True
        \STATE \textbf{break}
    \ENDIF
    \ENDFOR
\ENDFOR
\STATE \textbf{return} $m$
\end{algorithmic}
\end{algorithm}


Finally, with the obtained similarity matching result list $\mathcal{S}$, we compute the \textbf{Precision} ($P$) and \textbf{Recall} ($R$) metrics. Recall is defined as the ratio of correctly predicted actions to the total number of actions in the ground-truth, i.e., $R = m / |\mathcal{L}^g|$. Similarly, Precision is the ratio of correctly matched actions to the total number of predicted actions, calculated as $P = m / |\mathcal{L}^p|$. To gain deeper insight into the matching performance, we compute two sets of Precision and Recall metrics: one considering the full matching of all three action elements, and the other focusing solely on operation type. This allows us to assess both the overall accuracy and the accuracy specifically at the operation-type level.

\noindent \textbf{RPA-like Replay.} To validate the semantic comparison metrics, we conducted replay tests of predicted action sequences in the same environment as the original video recordings. Due to the lack of a replay environment for the \textsc{ActReal} dataset as it is collected elsewhere, our analysis focuses on the \textsc{ActOne} dataset. We employed a modified version of the VLM-based UI navigation tool~\cite{ufo}, which accepts an action sequence in natural language as input. Despite extensive customization efforts, current implementation supports only nine cases, primarily involving \texttt{click} and \texttt{type} actions.

The replay results \app{(summarized in Table~\ref{table:replay} in the Appendix) }reveal that out of the nine test cases, six were successfully replayed with the predicted action sequences. Two of the three unsuccessful cases exhibited lower Precision and Recall values, whereas the six successful cases all achieved one for both Precision and Recall. This validation experiment demonstrates that the Precision and Recall metrics derived from semantic matching are consistent with actual success rates of RPA replay, thus validating their effectiveness as indicators of action extraction performance.

\vspace{10pt}
\section{Experiment Results}
\label{section:results}

This section presents the experimental results of evaluating the DF and DiffF methods using the \textsc{ActOne} and \textsc{ActReal} datasets. We also undertake a thorough error analysis to identify potential root causes of failures. Additionally, we perform several ablation studies to gain deeper insights into the effectiveness of our proposed methods.

\subsection{Setup}
In our experiments, we implement both DF and DiffF approaches in \texttt{Python} and utilize two prominent series of VLMs: the GPT series (GPT-4o/4o-mini)~\cite{GPT-4o} and the Gemini series (Gemini1.5-Pro/Flash)~\cite{geminimodels}. Note that we intentionally select a large and small models for each series to study if the model size plays a crucial role in our task. For all VLMs, we set the temperature to 0 and use the default API settings for other parameters. Given that the GPT series permits a maximum of 10 images, we configure the window size to 10 frames, with an overlap of 5 frames in the DF approach. The frame sampling rate is set to one frame per second for \textsc{ActOne}, whereas \textsc{ActReal} employs a rate of 2 frames per second due to the faster pace of user actions in \textsc{ActReal}.

\subsection{Results for the \textsc{ActOne} Dataset}

\begin{table}[t]
  \centering
  \scriptsize
  \begin{tabular}{c | c | c  c  c  c}
  \hline
  \hline
  Method & Model & \makecell[c]{Recall\\(Operation)} & \makecell[c]{Precision\\(Operation)} & \makecell[c]{Recall\\(All)} & \makecell[c]{Precision\\(All)} \\
  \hline
  \multirow{4}{*}{\makecell[c]{DF}} & Gemini1.5-Pro & 0.71 & 0.73 & 0.49 & 0.51\\
  & Gemini1.5-Flash & 0.69 & 0.59 & 0.30 & 0.26 \\
  & GPT-4o & \textbf{0.83} & \textbf{0.81} & \textbf{0.71} & \textbf{0.68} \\
  & GPT-4o-mini & 0.63 & 0.33 & 0.38 & 0.17 \\
  \hline
  \multirow{4}{*}{\makecell[c]{DiffF}} & Gemini1.5-Pro & 0.75 & 0.48 & 0.45 & 0.24 \\
  & Gemini1.5-Flash & 0.74 & 0.37 & 0.54 & 0.27 \\
   & GPT-4o & \textbf{0.87} & \textbf{0.66} & \textbf{0.76} & \textbf{0.59} \\
   & GPT-4o-mini & 0.59 & 0.26 & 0.45 & 0.19 \\
  \hline
  \hline
  \end{tabular}
  \caption{Evaluation results for the \textsc{ActOne} dataset. }
  \label{tb:basic_dataset_res}
\end{table}

The evaluation results for \textsc{ActOne} is given in Table~\ref{tb:basic_dataset_res}. It includes the Precision and Recall metrics for the assessment of all three action elements (denoted as ``All'') and a restricted evaluation focusing solely on the operation type (denoted as ``Operation''). Key observations include:
\begin{itemize}[noitemsep, left=0pt]
    \item \emph{Model Comparison}: GPT-4o outperformed all other models across the four metrics for both DF and DiffF methods, followed by Gemini1.5-Pro. The Precision and Recall values, ranging from $0.6$ to $0.85$, highlight the potential of VLMs for extracting user actions from desktop recordings. Conversely, the smaller models, GPT-4o-mini and Gemini1.5-Flash, showed a marked decline in performance, underscoring the inherent difficulty of the task.
    \item \emph{DF vs. DiffF}: DF and DiffF exhibit comparable performance, though DiffF shows slightly lower Precision. This observation implies that incorporating explicit frame differences might not be essential for the current VLMs.
    \item \emph{Operation vs. All}: Performance degradation from evaluating only operation type to a full evaluation is more significant in smaller models than in larger ones.
\end{itemize}
We also study the breakdown analysis by case domain (i.e., operation type) for both methods\app{ with the results for GPT-4o and Gemini1.5-Pro given in Table~\ref{tb:basic_dataset_res_break_down} in the Appendix}. The results reveal significant performance variation across operation types for different models and methods, with no clear indication of which operation type is consistently easier to extract. This suggests that these operation types may present comparable levels of difficulty for current VLMs.

\subsection{Results for the \textsc{ActReal} Dataset}

The evaluation results for \textsc{ActReal} is depicted in Table \ref{tb:advanced_dataset_res}, with same column settings as \textsc{ActOne}. We observe that:
\begin{itemize}[noitemsep, left=0pt]
    \item \emph{\textsc{ActReal} vs. \textsc{ActOne}}: The performance of both methods declined on \textsc{ActReal} compared to \textsc{ActOne}, particularly in the ``All'' type metrics. This suggests a notable domain shift between the datasets, with \textsc{ActReal} presenting greater challenges for VLMs.
    \item \emph{DF vs. DiffF}: DF shows less decline compared to DiffF, suggesting that DF is better suited for real-world scenarios.
    \item \emph{Model Comparison}: For DF, GPT-4o outperforms all other models; as for DiffF, Gemini1.5-Pro has the best operation Recall and Precision, whereas GPT-4o has best overall Precision and Recall.
    \item \emph{Operation vs. All}: For both methods and all models, Precision and Recall under ``All'' conditions significantly decrease compared to ``Operation'' counterpart, unlike in \textsc{ActOne}. This suggests that extracting details and context is more challenging in \textsc{ActReal}. A closer examination of \textsc{ActReal} reveals that it includes a diverse range of UI elements that are difficult to describe unambiguously, even for humans. Additionally, there are more frequent screen changes unrelated to user actions, such as sudden pop-ups, which introduce additional noise into VLM inputs. Consequently, the task of capturing accurate details and context in \textsc{ActReal} is notably more complex.
    
\end{itemize}

\begin{table}[t]
  \centering
  \scriptsize
  \begin{tabular}{c | c | c  c  c  c}
  \hline
  \hline
  Method & Model & \makecell[c]{Recall\\(Operation)} & \makecell[c]{Precision\\(Operation)} & \makecell[c]{Recall\\(All)} & \makecell[c]{Precision\\(All)} \\
  \hline
  \multirow{4}{*}{\makecell[c]{DF}} & Gemini1.5-Pro & 0.73 & \textbf{0.72} & 0.37 & 0.32 \\
  & Gemini1.5-Flash & 0.77 & 0.39 & 0.47 & 0.22 \\
  & GPT-4o & \textbf{0.82} & 0.70 & \textbf{0.53} & \textbf{0.45} \\
  & GPT-4o-mini & 0.73 & 0.46 & 0.41 & 0.27 \\
  \hline
  \multirow{4}{*}{\makecell[c]{DiffF}} & Gemini1.5-Pro & \textbf{0.64} & \textbf{0.79} & 0.22 & 0.26  \\
   & Gemini1.5-Flash & 0.59 & 0.43 & 0.26 & 0.16 \\
    & GPT-4o & 0.38 & 0.78 & \textbf{0.27} & \textbf{0.54}\\
   & GPT-4o-mini & 0.30 & 0.59 & 0.13 & 0.25 \\
  \hline
  \hline
  \end{tabular}
  \caption{Evaluation results for the \textsc{ActReal} dataset.} 
  \label{tb:advanced_dataset_res}
\end{table}

\subsection{Error Analysis}

We proceed with a detailed analysis of the failure cases in the DF and DiffF methods. Specifically, we focus on cases exhibiting errors in Precision and Recall metrics when applying GPT-4o to the \textsc{ActOne} dataset. Upon examination, they can be categorized into the following four error types:
\begin{itemize}
    \item \textbf{Visual Hallucination}: VLMs sometimes generates hallucinated content when interpreting visual inputs. For instance, the Frame Difference Descriptor in DiffF may incorrectly detect a slight scroll bar movement when it appears or disappears, leading the downstream Action Proposer to falsely suggest a \texttt{scroll} action.
    \item \textbf{Visual Blindness}: VLMs occasionally fail to detect critical UI changes. For example, in a failed case involving the \texttt{drag} of a browser tab, the Frame Difference Descriptor in DiffF fails to capture the tab movement. 
    \item \textbf{Inadequate Reasoning}: VLMs may exhibit insufficient reasoning over contextual information. It is often observed in the Action Proposer/Corrector when inferring/correcting actions based on prior outputs. Inadequate reasoning typically involves the inability to apply or recognize domain knowledge. For instance, the Action Proposer in DiffF suggests an incorrect \texttt{click} based on a style change in a drop-down menu during mouse hover, which could have been avoided by considering the lack of menu expansion.
    \item \textbf{Poor Instruction-Following}: VLMs can fail to follow complex or lengthy instructions. For example, the Action Corrector may not update the \texttt{type} action details, even when explicitly prompted to use supplementary data.
\end{itemize}

\begin{table}[t]
  \centering
  \scriptsize
  \begin{tabular}{c | c  c  c  c }
  \hline
  \hline
  Method & \makecell[c]{Visual\\Hallucination} & \makecell[c]{Visual\\Blindness} & \makecell[c]{Inadequate\\Reasoning} & \makecell[c]{Poor\\Instruction-Following} \\
  \hline
  DF & 7 & 5 & 2 & 1 \\
  DiffF & 8 & 7 & 17 & 4 \\
  \hline
  \hline
  \end{tabular}
  \caption{Count of failed cases when applying GPT-4o to the \textsc{ActOne} dataset.}
  \label{tb:error_analysis}
\end{table}

Table~\ref{tb:error_analysis} summarizes the count of failed cases when applying GPT-4o to the \textsc{ActOne} dataset using both DF and DiffF. A single failure may encompass multiple error types, particularly in the DiffF method. The majority of failures in DF are due to visual issues, while DiffF's failures mainly stem from challenges in reasoning. This can be explained by the fact that DF relies heavily on visual capabilities to detect UI changes and determine actions. In contrast, DiffF struggles with inferring actions by reasoning over complex UI change descriptions, which often include irrelevant details. The results in Table~\ref{tb:error_analysis} suggest that enhancing the visual capabilities of VLMs would significantly improve DF performance. However, for DiffF, reasoning and instruction-following capabilities are equally essential.

\subsection{Ablation Study}


\noindent \textbf{Effectiveness of Action Corrector.}  We evaluated the Action Corrector's importance in DF and DiffF by removing it when using GPT-4o. The results (Row 2 and 3 in Table~\ref{tb:ablation}) show significant performance drops across almost all metrics, confirming its essential role in both methods.

\noindent \textbf{Impact of Sliding Window.} The sliding window in DF addresses GPT-series input limits, but Gemini-series models have a more relaxed limit, allowing us to test performance without it. As shown in Row 4 of Table~\ref{tb:ablation}, removing the sliding window reduces Recall but increases Precision. A similar trend occurs without window overlap (not shown), suggesting that without these mechanisms, VLMs produce fewer actions, missing some but reducing irrelevant ones.

\noindent \textbf{Role of Explicit UI Change Extraction.} Comparing DF and DiffF reveals that DF relies solely on VLMs' capabilities, while DiffF depends on explicit UI change descriptions. This raises the question of whether combining original frames with explicitly extracted UI changes could enhance performance. To explore this, we conduct two ablation studies.

In the first study, we add all video frames as extra visual input to the Action Proposer in DiffF. As shown in Row 5 of Table~\ref{tb:ablation}, while some metrics improve, the gains are marginal. In the second study, we introduce bounding boxes around changed regions in DF's input frames. The results (last row of Table~\ref{tb:ablation}) show no improvement and even some decline. Detailed analysis reveals that the decline in Precision can be attributed to an overemphasis on localized screen areas, neglecting the broader context. 

These studies suggest that augmenting VLM attention through explicit UI changes may misguide the model, leading to a narrow focus at the expense of critical global information. Given DF's strong performance on both \textsc{ActOne} and \textsc{ActReal} datasets, the most effective method for action extraction with current VLMs may be to rely on their intrinsic capabilities, as attempts to enhance them with explicit UI change extraction may introduce unnecessary confusion.

\begin{table}[t]
  \centering
  \scriptsize
  \begin{tabular}{c | c  c  c  c}
  \hline
  \hline
  \makecell[c]{Method + Model + Dataset\\(Variation)} & \makecell[c]{Recall\\(Operation)} & \makecell[c]{Precision\\(Operation)} & \makecell[c]{Recall\\(All)} & \makecell[c]{Precision\\(All)} \\
  \hline
  \makecell[c]{DF + GPT-4o + \textsc{AO}\\(w/o Action Corrector)} & \makecell[c]{0.83\\(0.76$\textcolor{green}{\downarrow}$)} & \makecell[c]{0.81\\(0.63$\textcolor{green}{\downarrow}$)} & \makecell[c]{0.71\\(0.48$\textcolor{green}{\downarrow}$)} & \makecell[c]{0.68\\(0.40$\textcolor{green}{\downarrow}$)} \\
  \hline
   \makecell[c]{DiffF + GPT-4o + \textsc{AO}\\(w/o Action Corrector)} & \makecell[c]{0.87\\(0.89$\textcolor{red}{\uparrow}$)} & \makecell[c]{0.66\\(0.46$\textcolor{green}{\downarrow}$)} & \makecell[c]{0.76\\(0.35$\textcolor{green}{\downarrow}$)} & \makecell[c]{0.59\\(0.21$\textcolor{green}{\downarrow}$)} \\
   \hline
   \makecell[c]{DF + Gemini1.5-Pro + \textsc{AR}\\(w/o Sliding Window)} & \makecell[c]{0.73\\(0.51$\textcolor{green}{\downarrow}$)} & \makecell[c]{0.72\\(0.91$\textcolor{red}{\uparrow}$)} & \makecell[c]{0.37\\(0.22$\textcolor{green}{\downarrow}$)} & \makecell[c]{0.32\\(0.36$\textcolor{red}{\uparrow}$)}\\
  \hline
   \makecell[c]{DiffF + GPT-4o + \textsc{AO}\\(add frames to Proposer)} & \makecell[c]{0.87\\(0.88$\textcolor{red}{\uparrow}$)} & \makecell[c]{0.66\\(0.69$\textcolor{red}{\uparrow}$)} & \makecell[c]{0.76\\(0.76)} & \makecell[c]{0.59\\(0.61$\textcolor{red}{\uparrow}$)} \\
   \hline
   \makecell[c]{DF + GPT-4o + \textsc{AO}\\(w/ region diff annotation)} & \makecell[c]{0.83\\(0.84$\textcolor{red}{\uparrow}$)} & \makecell[c]{0.81\\(0.61$\textcolor{green}{\downarrow}$)} & \makecell[c]{0.71\\(0.60$\textcolor{green}{\downarrow}$)} & \makecell[c]{0.68\\(0.42$\textcolor{green}{\downarrow}$)} \\
  \hline
  \hline
  \end{tabular}
  \caption{Ablation study results for several method variations on \textsc{ActOne} (AO) and \textsc{ActReal} (AR). Default method results are shown outside brackets (from Tables~\ref{tb:basic_dataset_res} and \ref{tb:advanced_dataset_res}), with corresponding variation results in brackets.}
  \label{tb:ablation}
\end{table}

\vspace{10pt}
\section{Discussion}

The ability of VLMs to extract user actions from desktop recordings opens up significant opportunities. One notable application is Robotic Process Automation, as evidenced by our metric validation test, showing that user actions extracted using our methods can be replayed through VLM-based UI automation tools. Additionally, this technology encourages the adoption of video as another modality for logging user activities (i.e., vlogs). For example, documenting best practices or preparing guidelines traditionally involves time-consuming written documentation. With VLMs, we now can generate guidelines more efficiently by recording demonstration videos from which VLMs can derive the necessary information. Furthermore, vlogs can be used for capturing personalized activities, with VLMs analyzing this data to identify individual interaction patterns and integrate them into other tools for ubiquitous personalization.

\vspace{10pt}
\section{Conclusions}
In this paper, we proposed two novel VLM-based methods for extracting user actions from desktop recordings: the Direct Frame-Based Approach (DF) and the Differential Frame-Based Approach (DiffF). Our evaluation shows that the DF approach is more effective, achieving higher accuracy in identifying actions. These methods have significant potential for applications in Robotic Process Automation, video-based tutorials and guideline generation, and user personalization. This work serves as a foundation for future research in desktop video action extraction, with opportunities for refining VLM capabilities and exploring broader applications.  


\bibliography{main}

\section{Appendix}
\subsection{Details of Methodology}
\noindent We provide further details regarding the methodology employed in this work. First, we demonstrate the implementation of the Frame Difference Localizer, as illustrated in the Figure \ref{fig:comparator-example}. Subsequently, we present an example output from the Frame Difference Descriptor, as depicted in the corresponding Figure \ref{fig:descriptor-output}.
\subsection{Details of Experimental Results}
\noindent We provide additional experimental details. We first present the case domain-level experimental results on the ACTONE dataset, as shown in the Table \ref{tb:basic_dataset_res_break_down}. The key observations of the results are as follows: For the Direct Frame-Based (DF) approach, from the perspective of assessing all three action elements (ALL), GPT-4o performs best in click-type cases, followed by scroll-type, and performs worst in drag-type cases. Gemini performs best in type-type cases, followed by click-type, and performs worst in select-type cases. Overall, for the DF approach, both models perform better in click, scroll, and type actions, but perform worse in drag and select actions.
For the Differential Frame-Based (Diff) approach, GPT-4o performs best in click-type cases, followed by select-type, but performs worst in scroll-type cases. Gemini performs best in scroll-type cases, while its performance in type and click actions is relatively weaker. Overall, for the Diff approach, GPT-4o and Gemini show different strengths: GPT-4o performs better in click and select actions but worse in scroll, whereas Gemini excels in scroll actions.

\noindent Then, we describe the replay results on selected samples from the ACTONE dataset, where we reproduced 9 cases from the dataset, with the results summarized in the Table \ref{table:replay}. In the result, two-thirds of the cases can be successfully reproduced.

\subsection{Prompt}
\noindent We also list all relevant prompts. First, we present the prompt for the action proposer in the DF method, as shown in the Table \ref{prompt:action-proposer-df-1}. Then, we provide the prompt for the corrector in the DF method, followed by the prompt of the Frame Difference Descriptor, as shown in the corresponding Table \ref{prompt:corrector-df-1} and \ref{prompt:frame-difference-descriptor-1}. Finally, we present the prompts for the Action Proposer and the Action Corrector, as summarized in the Table \ref{prompt:action-proposer-1} and Table \ref{prompt:action-corrector-1}.

\newpage
\onecolumn
\begin{appendices}

\section{}

\begin{figure}[htb!]
 \centering
  \includegraphics[width=\columnwidth]{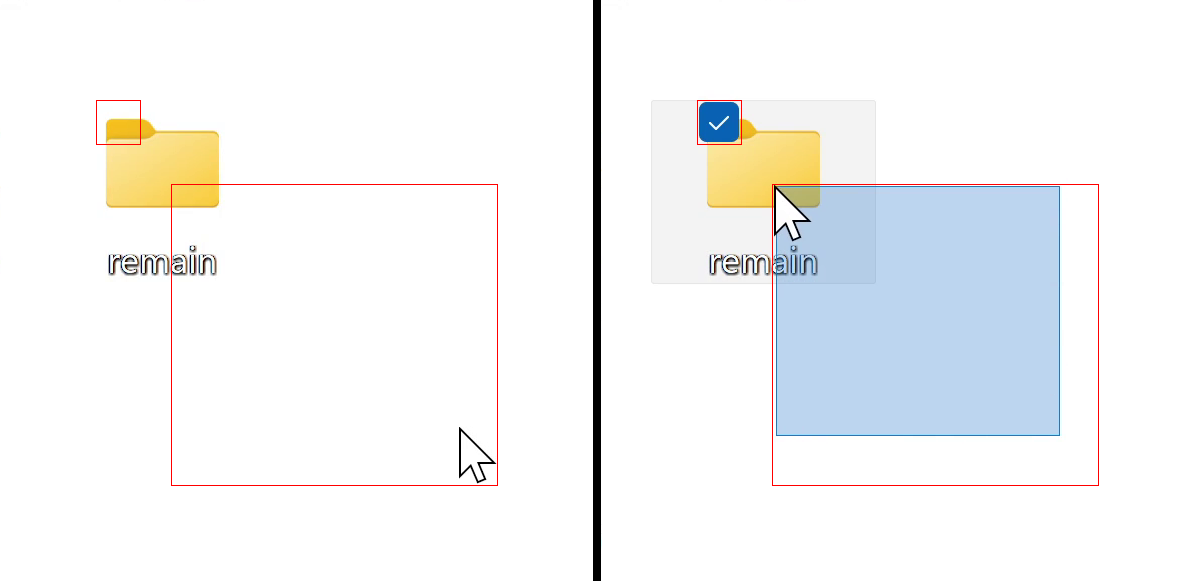}
  \caption{An example output from Frame Difference Localizer. Left and right are the changed regions with red bounding boxes in the previous and current frames, respectively.}
  \label{fig:comparator-example}
\end{figure}

\begin{figure}[H]
    \centering
    \begin{minipage}{0.9\textwidth}
        \centering
        \footnotesize
        \begin{Verbatim}
{
    "global_description": "The whole screenshot mainly contains a desktop with several icons, including Recycle Bin, a folder named 'demo', Google Chrome, and OBS Studio. The taskbar is visible at the bottom.",
    "description": "The region contains a folder named 'remain' and part of the desktop background. The folder is selected in the new image.",
    "changed": true,
    "old_cursor_shape": "normal",
    "new_cursor_shape": "normal",
    "changes": [
        {
            "subject": "folder",
            "type": "style_change",
            "old": "The folder 'remain' is not selected.",
            "new": "The folder 'remain' is selected with a blue checkmark and a blue selection box around it.",
            "message": "The folder 'remain' has been selected."
        }
    ],
    "frame": 2,
    "index": 0,
    "id": "2_0"
}
        \end{Verbatim}
    \end{minipage}
    \caption{An example output of Frame Difference Descriptor. It is derived from the identified UI changes in Figure~\ref{fig:comparator-example}.}
    \label{fig:descriptor-output}
\end{figure}

\newpage

\begin{table}[H]
  \centering
  \begin{tabular}{c | c | c | c | c | c | c}
  \hline
  \hline
  Method & Model & \makecell[c]{Case Domain\\(Video Count)} & \makecell[c]{Recall\\(Operation)} & \makecell[c]{Precision\\(Operation)} & \makecell[c]{Recall\\(All)} & \makecell[c]{Precision\\(All)} \\
  \hline
  \multirow{10}{*}{DF} & \multirow{5}{*}{GPT-4o} & \texttt{click} (14) &  0.94 & 0.88 & \textbf{0.90} & \textbf{0.81}\\
   & & \texttt{select} (11) &  \textbf{1.00} & \textbf{0.95} & 0.64 & 0.64\\
   & & \texttt{scroll} (6) &  0.75 & 0.75 & 0.75 & 0.75\\
   & & \texttt{drag} (5) &  0.40 & 0.50 & 0.30 & 0.40\\
   & & \texttt{type} (4) &  0.67 & 0.63 & 0.67 & 0.63\\
   \cline{2-7}
  & \multirow{5}{*}{Gemini1.5-Pro} & \texttt{click} (14) &  \textbf{0.85} & \textbf{0.87} & 0.63 & 0.65\\
   & & \texttt{select} (11) &  0.64 & 0.64 & 0.18 & 0.18\\
   & & \texttt{scroll} (6) &  0.58 & 0.67 & 0.58 & 0.67\\
   & & \texttt{drag} (5) &  0.70 & 0.63 & 0.50 & 0.43\\
   & & \texttt{type} (4) &  0.67 & 0.75 & \textbf{0.67} & \textbf{0.75}\\
   \hline
  \multirow{10}{*}{Diff} & \multirow{5}{*}{GPT-4o} & \texttt{click} (14) &  0.82 & 0.71 & 0.82 & \textbf{0.71}\\
   & & \texttt{select} (11) &  \textbf{1.00} & 0.77 & 0.82 & 0.64\\
   & & \texttt{scroll} (6) &  0.83 & 0.41 & 0.67 & 0.37\\
   & & \texttt{drag} (5) &  0.70 & 0.37 & 0.70 & 0.37\\
   & & \texttt{type} (4) &  0.92 & \textbf{0.88} & 0.58 & 0.63\\
   \cline{2-7}
  & \multirow{5}{*}{Gemini1.5-Pro} & \texttt{click} (14) &  0.69 & 0.49 & 0.36 & 0.18\\
   & & \texttt{select} (11) &  0.73 & 0.44 & 0.45 & 0.21\\
   & & \texttt{scroll} (6) &  \textbf{0.92} & \textbf{0.60} & \textbf{0.75} & \textbf{0.57}\\
   & & \texttt{drag} (5) &  0.80 & 0.33 & 0.60 & 0.22\\
   & & \texttt{type} (4) &  0.67 & 0.55 & 0.08 & 0.13\\
  \hline
  \hline
  \end{tabular}
  \caption{Evaluation results for the \textsc{ActOne} dataset categorized by case domain. }
  \label{tb:basic_dataset_res_break_down}
  \vspace{-4mm}
\end{table}

\begin{table}[H]
\centering
\begin{tabular}{c |c|c|c}
\hline
\hline
\multirow{2}{*}{Test Case in \textsc{ActOne}} & \multicolumn{2}{|c|}{Semantic Matching} & \multirow{2}{*}{Successful Replay?} \\
\cline{2-3}
 & Precision (All) & Recall (All) &  \\
\hline

\texttt{click/icon/taskbar} & 1 & 1 & yes \\ 
\hline
\texttt{click/text/checkbox} & 1 & 1 & yes \\ 
\hline
\texttt{click/text/dropdown} & 0.5 & 0.5 & no \\ 
\hline
\texttt{click/text/link} & 1 & 1 & yes \\ 
\hline
\texttt{click/text/text\_field} & 1 & 1 & yes \\ 
\hline
\texttt{click/text\_icon/menu} & 1 & 1 & yes \\ 
\hline
\texttt{click/text\_icon/tab} & 1 & 1 & no \\ 
\hline
\texttt{type/number} & 0.5 & 1 & no \\ 
\hline
\texttt{type/word} & 1 & 1 & yes \\ 
\hline
\hline
\end{tabular}

\caption{Replay results of test cases in \textsc{ActOne}. Precision and Recall metrics from semantic matching are shown for comparison.}
\label{table:replay}
\end{table}

\begin{table*}[htbp]
    \centering
    \footnotesize
    \begin{tabular}{p{15cm}}
    \toprule

\begin{Verbatim}[breaklines=true]
  - You are a screen recording video analysis agent.
  - Your task is to first identify the user's <Operation> from the provided <Video> and then output the full user's <Operation Sequence>.

  # <Video> description:
    - <Video> is sperated into screenshot frames. Each frame is a snapshot of the application interface, and the user may interact with the interface elements.
    - The computer environment can be Windows, Mac, or Linux operating systems.
    - The application consists of Office 365, desktop, web browser, and other applications.
    - The application interface element is composed of multiple elements, e.g., buttons, dropdowns, icons, text fields, and other interactive elements.
    - The user's <Operation> on the application interface element consists of click, drag, scroll, select, and type operations, e.g., click on a button, drag an icon, scroll a page, select text, or type in a text field.
  
  # <Operation> description and identification methods:
    ## click
      Description: 
        - The user clicks on an interface element(e.g., a button, link, or icon), activating the element(e.g., button press effect), and triggering various events(e.g., opening a new window, expanding a menu, changing the state of an element).
        - If you think the user's operation is "click", you need to keep observing several frames to see if the operation is "drag" or "select", which contains the "click" operation.
      Identification:
        - By the change of mouse:
          - shape change: The mouse may change shape (e.g., pointing hand when clicking a link).
        - By the change of interface element:
          - press effective: Buttons or other clickable elements display a press effect (e.g., changing color, showing a shadow, slightly changing shape, checkbox gets checked).
        - By the change of display:
          - feedback message: The interface displays feedback messages or changes(e.g., new window opens, menu expands).
    ## select
      Description: 
        - Text selection: The user selects text in a document, highlighting the selected text with a different background color. For example, select two sentences in Microsoft Word.
        - Icon selection: The user selects icons on a desktop or in an application, the selected icons should be enclosed within a blue rectangular box. For example, select two icons on the desktop.
        - Cell selection: The user selects cells in a spreadsheet or table, highlighting the selected cells with a different background color. For example, select three cells in Excel.
      Identification:
        - By the change of mouse:
          - Text selection:
            - mouse position: The position of the mouse indicates the start and end of the selection range. You should observe the mouse movement over the text to identify the selection.(e.g., from the beginning of the first sentence to the end of the second sentence)
          - Icon selection:
            - Mouse position: mouse move over the icons, and the selected icons should be enclosed within a blue rectangular box. You should observe the blue rectangular box region to identify the number of selected icons.
        - By the change of interface element:
          - color change: The background color of the selected text or selected icons changes to indicate the selection.(e.g., from white to blue)
    ## type
      Description: 
        - Text input: The user types text into a text field or document, entering characters, words, or sentences. For example, typing in a search bar, filling out a form, or writing an email.
        - Command input: The user types commands or inputs specific text strings to perform actions or trigger events. For example, typing commands in a terminal.
      Identification:
        - By the change of interface element:
          - text change: The content of the text field changes as the user types, updating the displayed text.
    ## scroll
      Description: 
        - Vertical scroll: The user scrolls vertically through a document or interface, moving the content up or down to view more information.
        - Horizontal scroll: The user scrolls horizontally through a document or interface, moving the content left or right to view more information.
      Identification:
        - By the change of mouse:
          - mouse position: The mouse may move to the scroll bar or scroll area, indicating the intention to scroll.
 
\end{Verbatim}
    \textbf{CONTINUE ON THE NEXT PAGE} \\
    \bottomrule
    \end{tabular}
    \caption{Prompt of action proposer of DF method: 1/3}
    \label{prompt:action-proposer-df-1}
\end{table*}

\begin{table*}[htbp]
    \centering
    \footnotesize
    \begin{tabular}{p{15cm}}
    \toprule

\begin{Verbatim}[breaklines=true]
        - By the change of interface element:
          - scroll bar movement: Observe the movement of the scroll bar. The vertical scroll bar is usually located on the right side of the interface, and the horizontal scroll bar is usually located at the bottom of the interface.
        - By the change of display:
          - Content movement: The content of the document or interface moves up or down (vertical scroll) or left or right (horizontal scroll) as the user scrolls.   
    ## drag
      Description:
        - Icon drag: The user drags an icon, moving it from one location to another within the interface. For example, dragging a file to a folder, moving an icon on the desktop.
        - Text drag: The user drags selected text, moving it to a different location within a document or interface. For example, rearranging paragraphs in a document.
      Identification:
        - By the change of position of the icon or text:
          - object position change: The object or text moves with the mouse, indicating the drag operation.
        - By the change of mouse or mouse:
          - Mouse position: The mouse moves with the dragged object or text.
        - By the change of interface element:
          - drag indicator: The object or text being dragged shows a visual indicator of movement, such as a shadow, outline, or placeholder.

  # The <Difference> between operations:
    - Drag vs. click: Drag operation include click operation, but drag operation involves moving an object from one location to another, while click operation involves triggering actions or changing the state of elements.
    - Drag vs. select: 
      - Drag operation involves moving an object from one location to another, while select operation involves highlighting text or selecting items, don't involve moving objects.
      - Icon selection is a type of select operation, not a drag operation.
    - Select vs. click: Select operation include click operation, but select operation involves highlighting text or selecting items, while click operation involves triggering actions or changing the state of elements.
    - Click vs. scroll: The click operation involves triggering actions or changing the state of elements, while the scroll operation involves moving the content up or down to view more information.

  # General guidelines for extracting <Operation Sequence>:
    - The <Operation Sequence> should be ordered by the timestamp of the operations, starting from the earliest operation to the latest operation.
    - For each frame, you should do as follows:
      - First focus on the position of the mouse and the elements near the mouse.
      - Identify the user's operation <Operation> and the associated information. 
      - If the operation is continuous(e.g., drag, select, scroll), you should observe several frames to identify the complete operation. E.g., if the user is selecting multiple sentences, you should observe the selection process until the user finishes the selection.
      - If the operation is compound, keep observing several frames to identify the complete operation.
      - If the operation is not clear, you can skip it.
    - For each operation, extract the following information:
      - "frame_total": The total number of frames in the video.
      - "frame_idx": The start and end frame index of the operation, e.g., [1, 3] means the operation is from the 1st frame to the 3rd frame.
      - "mouse_position": The position of the mouse or pointer on the screen, described in relation to the interface elements or text near it.
      - "element_state_pre_interaction": The state of the element before the interaction. If the operation is scroll, you should include the content before the scroll. If operation is drag, click, or select, you should include the state of the element near the mouse.
      - "element_state_after_interaction": The state of the element after the interaction. If the operation is scroll, you should include the content after the scroll. If the operation is drag, click,or select, you should include the state of the element near the mouse.
      - "Thoughts": Why you think the operation is the specific category . (e.g., "The mouse clicks the icon and move it to another location, which indicates a drag operation.")
      - "operation_category": The category of operation. 
        - It can only one of the following: "Click", "Drag", "Scroll", "Select", "Type".
        - Choose the one based on principle as follows:
          - Your choice should be based on thought, mouse position, element_state_pre_interaction, element_state_after_interaction.
          - You should distinguish between click, select, drag, and click operations according to the <Difference> between operations.
          - If the operation is "drag" or "select", you should choose "drag" or "select" instead of "click".
      - "application": The name of the application that the user is interacting with, including the application category and identifier.
        - category: The application category can include but not limit to one of the following:  Microsoft Word, Microsoft Excel, Microsoft PowerPoint, Web Browser, Windows OS, etc.
        - identifier: The application identifier should be a more specific description of the application  
\end{Verbatim}
    \textbf{CONTINUE ON THE NEXT PAGE} \\
    \bottomrule
    \end{tabular}
    \caption{Prompt of action proposer of DF method: 2/3}
    \label{prompt:action-proposer-df-2}
\end{table*}

\begin{table*}[htbp]
    \centering
    \footnotesize
    \begin{tabular}{p{15cm}}
    \toprule

\begin{Verbatim}[breaklines=true]
          - If category is "Web Browser", the identifier can be the name of the website (e.g., "Google", "YouTube", "Apple Music").
          - if category is "Windows OS", the identifier can be the name of the window or the desktop (e.g., "Desktop", "Taskbar", "File Explorer", "Menu").
          - If there is no specific identifier, leave it empty(e.g., "")."
      - "target_object": The object in "application" that the user interacted with, including the object category and identifier.
        - category: The object category can only be one of the following: "button", "text field", "icon", "dropdown", "list", "scroll bar", "document", "webpage", "dialog box", "menu", "file", "folder", "checkbox", "radio button", "search bar", "form", "email", "paragraph", "sentence", "word".
        - identifier: The identifier should be a description or name of the object (e.g., "Bold button", "Main Text Area", "File icon").
          - If there is no specific identifier, leave it empty(e.g., "")."
          - For the "scroll bar" category, the identifier should be the  direction of the scroll bar and the subject that the scroll bar control.(e.g. horizontal scroll bar of sheet1)
      - "additional_info": Any additional information related to the operation, such as the direction of scroll, the amount of scroll, the content typed, or the location of selected text or icons.
        - "additional_info" is optional and should be included only for the operation category "scroll", "type", and "select", for other categories, you can leave it empty(e.g., "").
        - For the "scroll" operation, include the direction of scroll ("up" or "down", "left" or "right") and the amount of scroll (e.g., "half page", "two lines", "until that you can see the yellow icon in the dropdown list").
        - For the "select" operation, include the content selected. 
          - For text selection, include the specific text selected. The granularity of the selection can be at the paragraph, sentence, or word level. e.g., "hello" in world level, "Hello World" in sentence level, "Hello World, How are you?" in paragraph level.
          - For icon selection, include the selected icons. The granularity of the selection can be at the icon level. e.g., "google icon", "apple icon".
          - For cell selection, include the selected cells. The granularity of the selection can be at the cell level. e.g., "A1", "B2".
        - For the "type" operation, include the content typed. 
          - If the user types character by character, concatenate multiple characters into one word.(e.g., concatenate "H", "e", "l", "l", "o" into "Hello").
          - If the word typed is too long, only output the number of sentences or the first few words. For example, "the first sentence", "the first three words". 
        - For the "drag" operation, include the initial and destination position of the object. For example, "from the right to left".
      - "abstract": The abstract description based on the above information, formatted as "operation_category" + "target_object" + "application" + "additional_info"(if needed), you should make it more fluent and readable.
  
  #  Output format of <Operation Sequence>:
    - The output should be in JSON format.
    - The JSON object should contain an array of user operations, each represented as a JSON object containing the extracted information for that operation.
    - You should avoid redundancy and repetition in the response.
    - Example output format:
      {
        "user_operations": [
          {
            "timestamp": "[1, 3]",
            "mouse_position": "near the text 'Hello World'",
            "element_state_pre_interaction": "Application window with text 'Hello World'",
            "element_state_after_interaction": "Application window with the text 'Hello World' selected",
            "thoughts": "The mouse is near the 'hello world' and the background the text changed",
            "operation_category": "select",
            "target_object": {
              "category": "text field",
              "identifier": "Main Text Area"
            },
            "application": {
              "category": "Microsoft Word",
              "identifier": ""
            },
            "additional_info": "Hello World",
            "abstract": "User selected 'Hello World' in the Main Text Area in Microsoft Word"
          }
        ]
      }
        
\end{Verbatim}
    \textbf{CONTINUE ON THE NEXT PAGE} \\
    \bottomrule
    \end{tabular}
    \caption{Prompt of action proposer of DF method: 3/3}
    \label{prompt:action-proposer-df-3}
\end{table*}

\begin{table*}[htbp]
    \centering
    \footnotesize
    \begin{tabular}{p{15cm}}
    \toprule

\begin{Verbatim}[breaklines=true]
  - You are a post-processing agent.
  - Your input is a sequence of user operations extracted from the interface images, which may contain errors or redundancies.
  - Your task is to post-process the operations according to the <GUIDELINE> of the post-processing.
  - You should output the chain of thoughts according to the <GUIDELINE> of the chain of thoughts before the final result.
  - Output the final result in a structured format according to the <FORMAT> of the output.

  # <GUIDELINE> of the post-processing
    ## Check the redundant operations
      - First, you should check the adjacent operations, and if the adjacent operations are the same operation, and their target_object is the same, you should keep the first operation and remove the redundant operation. E.g., if there are two "click" operations on the same button, you should keep the first "click" operation and remove the second "click" operation.
      - Second, you should check the adjacent operations, and if the one operation is the sub-operation of the other operation, you should remove the sub-operation. E.g., if there is a "click" operation followed by a "drag" operation, you should remove the "click" operation and keep the "drag" operation.
    ## Check the reasonableness of the operations
      - First, you should check the operation category according to Your thoughts. If the operation category is not reasonable, you should correct it.
      - Second, you should check the target_object according to the thoughts and abstract. If the target_object is not reasonable, you should correct it.
      - Third, you should check the application according to the thoughts and abstract. If the application is not reasonable, you should correct it.
    ## Check the completeness of the operations
      - First, you should check the additional_info according to the operation category. If the additional_info is missing, you should complete it.
      - Second, you should check the abstract according to the operation category, target_object, application, and additional_info. If the abstract is missing or not fluent, you should complete or correct it.
  
  # <GUIDELINE> of the chain of thoughts
    - First, check the redundant operations according to the <GUIDELINE> of the post-processing. And give the reason why you think the operation is redundant.
    - Second, check the reasonableness of the operations according to the <GUIDELINE> of the post-processing. And give the reason why you think the operation is not reasonable.
    - Last, check the completeness of the operations according to the <GUIDELINE> of the post-processing. And give the reason why you think the operation is not complete.

  # <GUIDELINE> of the output
  - The output should be in strictly valid JSON format, with no extra text or characters before or after the JSON.
  - If there are no user operations, you must return the 'user_operations' key with an empty list as its value.
    Example output format:
    {
      "user_operations": [
        {
          "thoughts": "The mouse is near the 'hello world' and the background of the text changed",
          "operation_category": "select",
          "target_object": {
            "category": "text field",
            "identifier": "Main Text Area"
          },
          "application": {
            "category": "Microsoft Word",
            "identifier": ""
          },
          "additional_info": "Hello World",
          "abstract": "User selected 'Hello World' in the Main Text Area in Microsoft Word"
        }
      ]
    }
    \end{Verbatim}
    \\
    \bottomrule
    \end{tabular}
    \caption{Prompt of corrector of DF method}
    \label{prompt:corrector-df-1}
\end{table*}

\begin{table*}[htbp]
    \centering
    \footnotesize
    \begin{tabular}{p{15cm}}
    \toprule

\begin{Verbatim}[breaklines=true]
  - You are an operation merge agent.
  - You will receive a list of user operations that are extracted from video frames using the sliding window method; the definition of the input is in <INPUT DETAIL>.
  - Your task is to delete the repeated operations caused by the overlapping of the adjacent windows and merge the entire operation sequence, you can refer to the <GUIDELINE FOR MERGING> for the merging rules.
  - You should output the merged operation sequence in the same format of <OUTPUT FORMAT>

  # <INPUT DETAIL>:
  - The input is a list of user operations extracted from the video frames.
  - Each item in the list is a JSON object containing the start and end frame of the sliding window,with a list of user operations extracted from the window.

  # <GUIDELINE FOR MERGING>:
  - For each pair of neighboring sliding windows, pay attention to actions at the end of the previous window and those at the beginning of the next window, check if certain actions match one of the following merging criteria:
    - Are there "drag" actions on the same element (or different element description referring to the same element) in the same app?
      - Write these actions down, which should be merged into one "drag" later.
    - Are there scroll actions on the same element (or different element descriptions referring to the same element) in the same app?
        - Write these actions down, which should be merged into one "scroll" later.
    - Are there "type" actions in the same app?
        - For such "type" actions, is earlier action "additional_info" (the typed text) a prefix of later action "additional_info"? If so, they actually belong to the same sequence of character typing and should be merged into one "type" action.
        - Write the actions satisfying the conditions, which should be merged to one "type" later.
    - Are there "select" actions in the same app?
        - For such "select" actions, is earlier action "additional_info" (selected items) a subset/superset of later action "additional_info"? If so, they actually belong to the same sequence of selecting and should be merged into one "select" action.
        - Write the actions satisfying the conditions, which should be merged to one "select" later.
    - Is there a "select" (or "click") action followed by a "drag" action on the same element (or different element description referring to the same element) in the same app?
        - Confirm if "select" is followed by "drag".
        - Write the actions satisfying the conditions, which should be merged into one "drag" later.
    - Is there a "select" (or "click") action followed by a "drag" action on same element (or different element description referring to the same element) in the same app?
        - Confirm if "select" is followed by "drag"".
        - Write the actions satisfying the conditions, which should be merged into one "drag" later.

  
  # <OUTPUT FORMAT>:
  - The output should be in JSON format.
  - The JSON object should contain an array of user operations, each represented as a JSON object containing the extracted information for that operation.
    Example output format:
    {
      "user_operations": [
        {
          "thoughts": "The mouse is near the 'hello world' and the background of the text changed",
          "operation_category": "select",
          "target_object": {
            "category": "text field",
            "identifier": "Main Text Area"
          },
          "application": {
            "category": "Microsoft Word",
            "identifier": ""
          },
          "additional_info": "Hello World",
          "abstract": "User selected 'Hello World' in the Main Text Area in Microsoft Word"
        }
      ]
    }

    \end{Verbatim}
    \\
    \bottomrule
    \end{tabular}
    \caption{Prompt of merger of DF method}
    \label{prompt:merger-df-1}
\end{table*}

\begin{table*}[htbp]
    \centering
    \footnotesize
    \begin{tabular}{p{15cm}}
    \toprule

\begin{Verbatim}[breaklines=true]
<<<CONTEXT>>>

A user is operating with a mouse and/or keyboard in Windows, which is recorded and requires analysis.

<<<INPUT>>>

The image input contains 2 images:
- one image is a full screenshot of the current screen, with a yellow box drawn around the region where UI changes might have happened.
- the other image represents the side-by-side comparison of the region bounded by yellow box before and after the likely UI change. The image is a pair of sub-images separated by a vertical black line, where left is the old looking of UI (1 second ago) and right is the new looking (now).
Note that the image may or may not contain UI changes.

Red boxes are drawn around changed pixels for you to better identify them.

The text input includes 2 fields for you to have a rough idea of the screen location of the region, and what the region would likely contain at this location:
- "screen_resolution": the screen resolution of the complete screenshot, in the format of [height, width].
- "bbox": the location and size of the yellow box in the screenshot provided to you, in the format of [minr, minc, maxr, maxc].

<<<TASK>>>
Your goal is to describe the differences between the old and new looking of UI in the comparison image (second one). You should focus on the second image while the first image is for you to understand the context.
You need to output a JSON object with the following fields:
- "global_description" (string): brief overall description of what is contained in the full screenshot (first image).
- "description" (string): brief description of what is contained in visible region (second image), and what UI changes have happened if any. If you recognize well-known icons or logos, you should also mention them here.
- "changed" (bool): indicating whether the UI has changed in the image.
- "old_cursor_shape" (string): The shape of the cursor in the old looking, in common terms like "normal", "crosshair", "I-beam" etc. Leave it null if cursor is not present in old image.
- "new_cursor_shape" (string): The shape of the cursor in the new looking, in common terms like "normal", "crosshair", "I-beam" etc. Leave it null if cursor is not present in new image.
- "changes" (list): a list of JSON objects, each describing a UI change. Leave it empty if "changed" is false. Each object should have the following fields:
    - "subject" (string): The subject of the UI change.
    - "type" (string): One of "appear", "disappear", "move", "rotate", "text_content_change", "style_change".
    - "old" (string): Description of the old looking of the subject in the image.
    - "new" (string): Description of the new looking of the subject in the image.

    - "message" (string): Explanation of what the UI change indicates. Do not suggest UI actions, but just what information the changes are intended to convey. Each change may have more than one explanations, use your critical thinking to list as many sound explanations as possible.
      Examples are: "the UI element is focused", "the UI element is defocused", "the UI element is selected", "The app has notifications", "a new window is open".
Note that there may be more than one UI change in the image. You should describe all the changes you can observe in the above fields.

Following are example outputs.

<<<EXAMPLE OUTPUT 1>>>
{
    "global_description": "The whole screenshot mainly contains an open Excel window, with a worksheet displayed.",
    "description": "The region contains an open Excel window, with a worksheet displayed.",
    "changed": true,
    "old_cursor_shape": null,
    "new_cursor_shape": null,
    "changes": [
        {
            "subject": "window",
            "type": "appear",
            "old": "the region is part of the desktop background",
            "new": "the region contains an open Excel window",
            "message": "The Excel window has been opened or moved to the region, or changed from minimized to opened state"
        }
    ]
}
\end{Verbatim}
    \textbf{CONTINUE ON THE NEXT PAGE} \\
    \bottomrule
    \end{tabular}
    \caption{Prompt of Frame Difference Descriptor : 1/2}
    \label{prompt:frame-difference-descriptor-1}
\end{table*}

\begin{table*}[htbp]
    \centering
    \footnotesize
    \begin{tabular}{p{15cm}}
    \toprule

\begin{Verbatim}[breaklines=true]
<<<EXAMPLE OUTPUT 2>>>
{
    "global_description": "The whole screenshot mainly contains a Word document, with a paragraph of text displayed.",
    "description": "The region contains part of the blank area of the document.",
    "changed": true,
    "old_cursor_shape": null,
    "new_cursor_shape": "normal",
    "changes": [
        {
            "subject": "cursor",
            "type": "move",
            "old": "the cursor is at the left of the region",
            "new": "the cursor is at the right of the region",
            "message": "The cursor has been moved from left to right"
        }
    ]
}
<<<EXAMPLE OUTPUT 3>>>
{
    "global_description": "The whole screenshot mainly contains a desktop with a few icons displayed.",
    "description": "The region contains part of the desktop background without any icons.",
    "changed": true,
    "old_cursor_shape": "I-beam",
    "new_cursor_shape": null,
    "changes": [
        {
            "subject": "cursor",
            "type": "disappear",
            "old": "the cursor is present in the region",
            "new": "the cursor is absent in the region",
            "message": "The cursor has been hidden or moved out of the region"
        }
    ]
}
<<<EXAMPLE OUTPUT 4>>>
{
    "global_description": "The whole screenshot mainly contains a browser window, with a search bar displayed.",
    "description": "The region contains a search bar with the text 'hello world' displayed.",
    "changed": true,
    "old_cursor_shape": "I-beam",
    "new_cursor_shape": null,
    "changes": [
        {
            "subject": "text",
            "type": "text_content_change",
            "old": "the text in the visible area is 'hello'",
            "new": "the text in the visible area is 'hello world'",
            "message": "More text has been added to the input field"
        }
    ]
}
<<<EXAMPLE OUTPUT 5>>>
{
    "global_description": "The whole screenshot mainly contains a text editor window, with some text displayed.",
    "description": "The region contains a line of text in yellow.",
    "changed": true,
    "old_cursor_shape": null,
    "new_cursor_shape": null,
    "changes": [
        {
            "subject": "text",
            "type": "style_change",
            "old": "the text in the region is black",
            "new": "the text in the region is yellow",
            "message": "the text color has changed"
        }
    ]
}
<<<EXAMPLE OUTPUT 6>>>
{
    "changed": false,
    "changes: []
}
    \end{Verbatim}
    \\
    \bottomrule
    \end{tabular}
    \caption{Prompt of Frame Difference Descriptor : 2/2 (continued)}
    \label{prompt:frame-difference-descriptor-2}
\end{table*}

\begin{table*}[htbp]
    \centering
    \footnotesize
    \begin{tabular}{p{15cm}}
    \toprule

\begin{Verbatim}[breaklines=true]
<<<CONTEXT>>>

A user is operating with mouse and/or keyboard in Windows, which is recorded and requires analysis.

<<<TASK>>>

Note that due to random processing errors, some of the input events may be inaccurate or even contradictory, hence it may be impossible to generate a list of actions coherent to all input events.
Your goal, in this case, is to aggregate the input events into a list of coherent and most likely actions. Specifically, you should prefer short list of actions, each of which explains many of events, rather than a long list of
actions, each of which explains only 1 or 2 events.

<<<INPUT>>>

The user input is a list of JSON objects, each representing a UI change in a region of the screen.

<<<OUTPUT>>>

Your output should be a list of JSON objects, each representing a coherent and most likely action. Each object should have the following fields:
- "app" (string): The name of the application where the actions are performed.
- "element" (string): The UI element where the actions are performed.
- "action" (string): The action performed on the UI element. One of "click", "drag", "scroll", "select", "type".
- "region" (string): The id of the region in which the UI change happened. For mouse actions, it should be the region where the cursor locates; for keyboard actions, it should be where the text changes happen. The id is in the format "<frame>_<index>"
- "evidences" (list): The list of (id string, reason string) paris of the input events that support this action. Each id is in the format "<frame>_<index>". The "region" id should be included in this list for consistency. Each evidence can only be used at most once in the output.


  # <Operation> description and identification methods:
    ## click
      Description: 
        - The user clicks on an interface element(e.g., a button, link, or icon), activating the element(e.g., button press effect) and triggering various events(e.g., opening a new window, expanding a menu, changing the state of an element).
        - If you think user's operation is "click", you need to keep observing several frames to see if the operation is "drag" or "select", which contains the "click" operation.
      Identification:
        - By the change of mouse:
          - shape change: The mouse may change shape (e.g., pointing hand when clicking a link).
        - By the change of interface element:
          - press effective: Buttons or other clickable elements display a press effect (e.g., changing color, showing a shadow, slightly changing shape, checkbox gets checked).
        - By the change of display:
          - feedback message: The interface displays feedback messages or changes(e.g., new window opens, menu expands).
    ## select
      Description: 
        - Text selection: The user selects text in a document, highlighting the selected text with a different background color. For example, select two sentences in Microsoft Word.
        - Icon selection: The user selects icons on a desktop or in an application, the selected icons should be enclosed within a blue rectangular box. For example, select two icons on the desktop.
        - Cell selection: The user selects cells in a spreadsheet or table, highlighting the selected cells with a different background color. For example, select three cells in Excel.
      Identification:
        - By the change of mouse:
          - Text selection:
            - mouse position: The position of the mouse indicates the start and end of the selection range. You should observe the mouse movement over the text to identify the selection.(e.g., from the beginning of the first sentence to the end of the second sentence)
          - Icon selection:
            - mouse position: mouse move over the icons, and the selected icons should be enclosed within a blue rectangular box. You should observe the blue rectangular box region to identify the number of selected icons.
        - By the change of interface element:
          - color change: The background color of the selected text or selected icons changes to indicate the selection.(e.g., from white to blue)
\end{Verbatim}
    \textbf{CONTINUE ON THE NEXT PAGE} \\
    \bottomrule
    \end{tabular}
    \caption{Prompt of Action Proposer : 1/3}
    \label{prompt:action-proposer-1}
\end{table*}

\begin{table*}[htbp]
    \centering
    \footnotesize
    \begin{tabular}{p{15cm}}
    \toprule

\begin{Verbatim}[breaklines=true]
    ## type
      Description: 
        - Text input: The user types text into a text field or document, entering characters, words, or sentences. For example, typing in a search bar, filling out a form, or writing an email.
        - Command input: The user types commands or inputs specific text strings to perform actions or trigger events. For example, typing commands in a terminal.
      Identification:
        - By the change of interface element:
          - text change: The content of the text field changes as the user types, updating the displayed text.
    ## scroll
      Description: 
        - Vertical scroll: The user scrolls vertically through a document or interface, moving the content up or down to view more information.
        - Horizontal scroll: The user scrolls horizontally through a document or interface, moving the content left or right to view more information.
      Identification:
        - By the change of mouse:
          - mouse position: The mouse may move to the scroll bar or scroll area, indicating the intention to scroll.
        - By the change of interface element:
          - scroll bar movement: Observe the movement of the scroll bar. The vertical scroll bar is usually located on the right side of the interface, and the horizontal scroll bar is usually located at the bottom of the interface.
        - By the change of display:
          - content movement: The content of the document or interface moves up or down (vertical scroll) or left or right (horizontal scroll) as the user scrolls.
    ## drag
      Description:
        - Icon drag: The user drags an icon, moving it from one location to another within the interface. For example, dragging a file to a folder, moving an icon on the desktop.
        - Text drag: The user drags selected text, moving it to a different location within a document or interface. For example, rearranging paragraphs in a document.
      Identification:
        - By the change of position of the icon or text:
          - object position change: The object or text moves with the mouse, indicating the drag operation.
        - By the change of mouse or mouse:
          - mouse position: The mouse moves with the dragged object or text.
        - By the change of interface element:
          - drag indicator: The object or text being dragged shows a visual indicator of movement, such as a shadow, outline, or placeholder.

  # The <Difference> between operations:
    - Drag vs. click: Drag operation include click operation, but drag operation involves moving an object from one location to another, while click operation involves triggering actions or changing the state of elements.
    - Drag vs. select: 
      - Drag operation involves moving an object from one location to another, while select operation involves highlighting text or selecting items, don't involve moving objects.
      - Icon selection is a type of select operation, not a drag operation.
    - Select vs. click: Select operation include click operation, but select operation involves highlighting text or selecting items, while click operation involves triggering actions or changing the state of elements.
    - Click vs. scroll: Click operation involves triggering actions or changing the state of elements, while scroll operation involves moving the content up or down to view more information.

In following section is an example output. Note that these events are unlikely to happen in the real world, but they are as such merely to illustrate the format of the output.

Remember:
    - all actions are one of the 5 types: click, drag, scroll, select, type. For 'select' action, make sure to name all the UI items selected.
    - The output should be in JSON format. Output one single JSON object instead of multiple ones.
    - The JSON object should contain an array of user operations, each represented as a JSON object containing the extracted information for that operation.
    - There may be multiple click actions in the recording. In this case, you should output multiple click actions, each with its own evidences. Keep in mind that the UI changes associated with a click happen in general within a few frames after the click event itself, so your evidences should not span over too many different frames.
    - If you find a "click" action, prefer "smaller" UI elements for the "element" field like "button", "cell", "option", avoid too vague descriptions, e.g. say "'refresh' option in context menu" instead of "context menu".
    - for "select", "click", "drag: actions, cursor must be in the new looking in the region where it happens.

\end{Verbatim}
    \textbf{CONTINUE ON THE NEXT PAGE} \\
    \bottomrule
    \end{tabular}
    \caption{Prompt of Action Proposer : 2/3 (continued)}
    \label{prompt:action-proposer-2}
\end{table*}

\begin{table*}[htbp]
    \centering
    \footnotesize
    \begin{tabular}{p{15cm}}
    \toprule

\begin{Verbatim}[breaklines=true]
    - "drag" and "scroll" actions must happen and be supported by regions where "move" events happen ("move" of anything but the cursor). To distinguish between "drag" and "scroll": if cursor is present, then cursor moving with ui element in motion indicates drag, and still cursor indicates scroll; If no cursor, then it's a scroll
    - For "drag" to happen, cursor must be in the new looking in the region where it happens.
    - "type" action can happen only if there is a at least one "text_change" in the region where it happens. On the other hand, text change can also happen without "type" action, e.g. caused indirectly by a click/select action else where.
    - If a scrollbar appears, it does not necessarily mean that a scroll action has happened. It may be a result of a click or drag action. An area of moving ui elements is a necessary condition of scroll action.

<<<EXAMPLE OUTPUT>>>
[
    {
        "app": "Microsoft Excel",
        "element": "cell A1",
        "action": "click",
        "region": "1_0",
        "evidences": [
            ["1_0", "cursor is in this region, and a bounding box appears around cell A1"],
            ["1_1", "The row A is highlighted"]
            ["1_2", "The column 1 is highlighted"]
            ["1_3", "The cell reference changed to A1"]
        ]
    },
    {
        "app": "Microsoft Excel",
        "element": "cell A2",
        "action": "type",
        "region": "3_1",
        "evidences": [
            ["3_1", "cursor is in this region, and text in it changed"],
            ["3_2", "text in this region changed"]
        ]
    }
    {
        "app": "Microsoft Word",
        "element": "main text area",
        "action": "type",
        "region": "5_1",
        "evidences": [
            ["5_1", "text changed from 'Hello' to 'Hello, World!'"]
        ]
    }
]
    \end{Verbatim}
    \\
    \bottomrule
    \end{tabular}
    \caption{Prompt of Action Proposer : 3/3 (continued)}
    \label{prompt:action-proposer-3}
\end{table*}

\begin{table*}[htbp]
    \centering
    \footnotesize
    \begin{tabular}{p{15cm}}
    \toprule

\begin{Verbatim}[breaklines=true]
Now you have to perform a sequence of tasks one by one, to improve your prediction.

The input of <TASK 1> is the JSON object containing the action sequence you just wrote. Each of the successive tasks takes the output of the previous task as input.
Make sure to follow the instructions of each task, by writing down your thoughts, writing the output, before going to the next task. Do not skip the thinking process or writing output of any of the tasks.

The tasks are to be performed in the following order:
<TASK 1>: correct "action" field
<TASK 2>: correct "app" field
<TASK 3>: correct "element" field for "select" actions
<TASK 4>: correct "element" field for "scroll" actions
<TASK 5>: correct "element" field for "type" actions
<TASK 6>: correct "element" field for "drag" actions
<TASK 7>: correct the "click" actions by analyzing their "evidences"
<TASK 8>: merge actions

Instructions of the tasks:
<TASK 1>: correct "action" field
Now you have to revise the "action" field of the action triples you just wrote, and correct all the actions that are not one of the five types ['click', 'type', 'scroll', 'select', 'drag']. This is usually because you used mistakenly one term from the events in your initial input. In this case, you should pick one of the above 5 verbs that best describes this action. For example:
    - If your wrong action is 'move', think twice: what UI elements were moved? Is it caused by a 'drag' or 'scroll'? The other 3 actions would unlikely to cause a move in common UI interactions.
    - If your wrong action is 'hover': recall that while hover may indeed have happened, but you should not output it as an action. Since valid cursor actions ('click', 'select', 'drag') are always followed by a hover, think step by step which of the 3 actions actually happened, based on the current region and evideces you are trying to explain in this action.

Also don't duplicate actions, that is, output the same action on same region twice. If so, you should remove the duplicates.
Remember to keep the actions without error intact. Fix the errors in the actions.
To complete your task, first write down your thoughts on how to correct these errors. Then, generate valid and complete JSON object that contains the actions with "action" field corrected, and all correct actions intact.

<TASK 2>: correct "app" field
    Now you have to revise the "app" field of the action triples you just wrote, and 

    - Avoid writing the "app" field too vaguely like "Windows". Instead, mention it in format of "<componenet> of Windows" for clarity. Examples are:
        - If the action happens on the desktop, output the "app" field as "Desktop of Windows".
        - If the action happens on the taskbar, output the "app" field as "Taskbar of Windows".
        - If the action happens on the start menu, output the "app" field as "Menu of Windows".
    - If the action happens on a web browser, output the "app" field as "<visible tab title> of Web Browser". For example, "Bing of Web Browser" or "The New York Times of Web Browser". One special case, is when the tab is a google search result, you should output the "app" field as "Google in Web Browser".

    Rememeber to leave the correct "app" values and all other fields intact in your output JSON object.
    If no correction is needed, simply output your prvious JSON object EXACTLY as it is, so that the downstream system can notice that no correction is made.

    To complete your task, first write down your step-by-step thoughts on how to correct the "app" fields if they have these issues.
    If one "app" field needs correction, but the current action item does not include enough information to correct it, make sure to go back to its supporting UI change events (find them by ids indicated in "evidences" and "region" fields) to review the original UI change events and correct the "app" field accordingly.
    After your thinking, output the complete and valid JSON object of actions with each of the "app" fields corrected.

<TASK 3>: correct "element" field for "select" actions
    Now you have to revise the "element" field of the action triples you just wrote. Consider the "select" action triples only, leaving all other actions intact.
    - If some text is selected, the "element" field should contain the selected text in single quotes (WITHOUT extra explanatory words like "text"), instead of the UI element including the text like "textbox".

        Rememeber to leave the correct "element" values and all other fields intact in your output JSON object.
        If no correction is needed, simply output your prvious JSON object EXACTLY as it is, so that the downstream system can notice that no correction is made.
        
\end{Verbatim}
    \textbf{CONTINUE ON THE NEXT PAGE} \\
    \bottomrule
    \end{tabular}
    \caption{Prompt of Action Corrector : 1/4}
    \label{prompt:action-corrector-1}
\end{table*}

\begin{table*}[htbp]
    \centering
    \footnotesize
    \begin{tabular}{p{15cm}}
    \toprule

\begin{Verbatim}[breaklines=true]
        To complete your task, first write down your step-by-step thoughts on how to correct/remove the "element" fields of the "select" actions if they have these issues.
        If one "element" field needs correction, but the current action item does not include enough information to correct it, make sure to go back to its supporting UI change events (find them by ids indicated in "evidences" and "region" fields) to review the original UI change events and correct the "element" field accordingly.
        After your thinking, output the complete and valid JSON object of actions with each of the "element" fields corrected.

<TASK 4>: correct "element" field for "scroll" actions
    Now you have to revise the "element" field of the action triples you just wrote. Consider the "scroll" action triples only, leaving all other actions intact.
    - For each scroll action in a web browser tab, output "element" as "<vertical/horizontal> scroll bar of <visible tab title>". Confirm whether the scroll bar is vertical or horizontal by looking at the UI change events that support the scroll action.
    - For every other scroll action, output "element" as "<vertical/horizontal> scroll bar of <UI element being scrolled>". Confirm whether the scroll bar is vertical or horizontal by looking at the UI change events that support the scroll action.

        Rememeber to leave the correct "element" values and all other fields intact in your output JSON object.
        If no correction is needed, simply output your prvious JSON object EXACTLY as it is, so that the downstream system can notice that no correction is made.

        To complete your task, first write down your step-by-step thoughts on how to correct/remove the "element" fields of the "scroll" actions if they have these issues.
        If one "element" field needs correction, but the current action item does not include enough information to correct it, make sure to go back to its supporting UI change events (find them by ids indicated in "evidences" and "region" fields) to review the original UI change events and correct the "element" field accordingly.
        After your thinking, output the complete and valid JSON object of actions with each of the "element" fields corrected.

<TASK 5>: correct "element" field for "type" actions
    Now you have to revise the "element" field of the action triples you just wrote. Consider the "type" action triples only, leaving all other actions intact. The typed content might be either text or other content:
    - If text was typed:
        - the "element" field should contain the typed text in single quotes (WITHOUT extra explanatory words like "text"), instead of the UI element like "textbox";
        - Make sure to summarize the complete typed text, by reviewing the "region" and "evidences" events that support the "type" action;
        - It may happen that the text you found from the UI events are still incomplete due to missing keyframes, so you should also take a look of the regions AFTER the "region" and "evidences" events (region ids are in form of "<frame>_<index>", so you should look at regions of greater frame numbers, not smaller) to infer the complete typed text. If such later regions exist, add the region ids to the "evidences" field of the current action.
    - If Something else is typed, observe among the the events: what content has appeared before I-beam cursor locations in the changed regions close-in-time to the "region" and "evidences" regions (that is, frame id should not differ with more than 2). Describe the new content briefly in the "element" field (without quotes). If new content has appeared in a UI event region, add its id to the "evidences" field of the current action.
    
        Rememeber to leave the correct "element" values and all other fields (except when you need to add more region ids to "evidences" of a "type" action) intact in your output JSON object.
        If no correction is needed, simply output your prvious JSON object EXACTLY as it is, so that the downstream system can notice that no correction is made.

        To complete your task, first write down your step-by-step thoughts on how to correct/remove the "element" fields of the "type" actions if they have these issues.
        If one "element" field needs correction, but the current action item does not include enough information to correct it, make sure to go back to its supporting UI change events (find them by ids indicated in "evidences" and "region" fields) to review the original UI change events and correct the "element" field accordingly.
        After your thinking, output the complete and valid JSON object of actions with each of the "element" fields corrected.

<TASK 6>: correct "element" field for "drag" actions
    Now you have to revise the "element" field of the action triples you just wrote. Consider the "drag" action triples only, leaving all other actions intact. The typed content might be either text or other content:
\end{Verbatim}
    \textbf{CONTINUE ON THE NEXT PAGE} \\
    \bottomrule
    \end{tabular}
    \caption{Prompt of Action Corrector : 2/4 (continued)}
    \label{prompt:action-corrector-2}
\end{table*}

\begin{table*}[htbp]
    \centering
    \footnotesize
    \begin{tabular}{p{15cm}}
    \toprule

\begin{Verbatim}[breaklines=true]
    - If items were dragged:
        - Does the "app" in which the drag happend allow dragging multiple items? If so, one or more items may be dragged; otherwise, only one item may be dragged. Some common apps:
            - Desktop: allow dragging multiple selected icons.
            - Taskbar: does not allow dragging multiple icons.
            - File Explorer: allow dragging multiple selected files.
            - Microsoft Word: allow dragging multiple selected lines, which must be contiguous.
            - List (in general): if items are selected, they can be dragged at the same time.
        - Name the items(s) being dragged specifically in "element", avoiding vague descriptions like "icons" or "items".
        - To identify the items being dragged, revisit the "region" and "evidences" events that support the "drag" action. In particular, what items kept moving in every supporting "region" and "evidences"? Were they moving with the cursor? During drag, some elements may move when their place was occupied by the dragged item. These items did not keep in all events following the cursor, and their movement was not caused directly by the drag, therefore they should not be included as dragged items.
    - If text was dragged:
        - the "element" field should contain the dragged text in single quotes (WITHOUT extra explanatory words like "text"), instead of the UI element like "textbox";
        - Observe selected text during the drag action, as text must be in selected state to be dragged;
        - Revisit "text_content_change" events that happened between start and end of drag actions. Does the selected text appear or disappear in these events? If so, it is likely that the text was dragged and moved, instead of being changed. The reason is that due to low frame sampling rate (1FPS), dragged/moved text may look like changed, while the intermediate dragging animation may be missing in low FPS keyframes.

        Rememeber to leave the correct "element" values and all other fields (except when you need to add more region ids to "evidences" of a "drag" action) intact in your output JSON object.
        If no correction is needed, simply output your prvious JSON object EXACTLY as it is, so that the downstream system can notice that no correction is made.

        To complete your task, first write down your step-by-step thoughts on how to correct/remove the "element" fields of the "drag" actions if they have these issues.
        Your thoughts must include the following steps:
        - Find out the "app" in which the drag happend. Does the app allow dragging multiple items? If so, one or more items may be dragged; otherwise, only one item may be dragged.
        - In "region" and "evidences" events that support the "drag" action, which items have moved? List them.
        - Among these items, which items kept moving in every supporting "region" and "evidences"? Were they moving with the cursor? During drag, some elements may move when their place was occupied by the dragged item. These items did not keep in all events following the cursor, and their movement was not caused directly by the drag, therefore they should not be included as dragged items.
        - List the items that kept moving in all supporting events with the cursor.
        - Review all UI events: in which event has one of the above specifaclly been selected? List them here for latet steps
        - If you found multiple items in the previous step, review the supporting events in "region" and "evidences", to see if they were moving in a translation manner, that is, in parallel. Items that have exchanged location or reordered cannot possibly have been dragged at the same time (since it is NOT a translation motion), and hence should not be included in the "element" field. List the items that were BOTH selected and moving in a translation manner.
        - If you found multiple items in the previous step, list all of them here for later output of "element" IF the app allows dragging multiple items; otherwise, list only one item here, which seems to be the most likely dragged item.

        If one "element" field needs correction, but the current action item does not include enough information to correct it, make sure to go back to its supporting UI change events (find them by ids indicated in "evidences" and "region" fields) to review the original UI change events and correct the "element" field accordingly.
        After your thinking, output the complete and valid JSON object of actions with each of the "element" fields corrected.


<TASK 7>: correct the "click" actions by analyzing their "evidences"
    Now you have to revise the "evidences" field of the "click" action triples you just wrote. Consider the "click" action triples only, leaving all other actions intact.

    Rememeber to leave the correct "evidences" values and all other fields intact in your output JSON object.
    If no correction is needed, simply output your prvious JSON object EXACTLY as it is, so that the downstream system can notice that no correction is made.

    To complete your task, first write down your step-by-step thoughts for each "click" action:
    - Review all the supporting UI events in the "evidences". If all evidences indicate that the curosor just happened to move over/beside the element, and no style change of the action elementhappend at all 
\end{Verbatim}
    \textbf{CONTINUE ON THE NEXT PAGE} \\
    \bottomrule
    \end{tabular}
    \caption{Prompt of Action Corrector : 3/4 (continued)}
    \label{prompt:action-corrector-3}
\end{table*}

\begin{table*}[htbp]
    \centering
    \footnotesize
    \begin{tabular}{p{15cm}}
    \toprule

\begin{Verbatim}[breaklines=true]
    - Write down what is expected to happen if this click action was actually performed and Walk through ALL UI events after the frame where the click supposedly happend (not only the evidences of the action) to check if the expected UI change actually happened. For example:
        - Clicking on a taskbar icon should open the corresponding app window, or display it if already open.
        - Clicking on a dropdown menu should open the dropdown menu, or close it if it was already open.
    - Distinguish between "click" and "hover": Some style changes happen on "hover", and UI events woul have been different than if it was a "click". For example, most slight text/background color change would likely by caused 
    - Indicator of "hover": the cursor moved over the element and then quickly moved out. After moving out, the element's looking changes back to the original state (the looking before cursor coming acrossing the element). Make sure to check a few frames before and after the evidence of this action to see if it's the case 
    - After the above steps, write down the updated list of evidences for this action, taking your above thoughts into account. Keep the evidences that support the "click" action, and remove the evidences that indeicate a "hover" action or another action type or no action at all.

    After your thinking, output the complete and valid JSON object of actions with each of "click" action corrected and all other actions intact. For each "click" action, if the new "evidences" list is non-empty, then keep the action item with updated "evidences" and all other fileds unchanged. If the new "evidences" list is empty, then remove the action item from the list of actions.

<TASK 8>: merge actions
    Now you have to revise the actions of type "type", "select", "drag", "click" you just wrote to see any of the same type actions are mergeable, leaving all the other actions intact.
    Indenfity all groups of actions that should have been merged, output one single merged action where:
    - "action" is the common action type of the actions being merged
    - "element" is a summary of "element" fields of all actions being merged, which contains all information of each individual "element". If "element" is in form of a set (e.g. set of icons, text as set of characters), use the "element" of the latest action (latest means greatest frame id)
    - "app" is the common app shared by all actions being merged
    - "evidences" is the union of "evidences" fields of all actions being merged
    - "region" is the region of the latest action (latest means greatest frame id)
    - all other fields are the same as the latest action (latest means greatest frame id)
    Indicators of actions of same type:
    - The actions happen in same app (necessary but not sufficient condition)
    - The actions have overlapping evidences or evidences that interleave or are close in time (in terms of id; evidences close in time must differ at most by 2 in frame id)
    - For "type" actions, ealier action "element" (in terms of frame id) is a prefix of later action "element", since text is typed in a sequence.
    - For "select" actions, earlier action "element" (in terms of frame id) is a subset OR superset of later action "element", since more or less text/items are selected during these frames

    Rememeber to leave the correct actions intact in your output JSON object.
    If no merging is needed, simply output your prvious JSON object EXACTLY as it is, so that the downstream system can notice that no correction is made.

    To complete your task, first write down your step-by-step thoughts on how to merge the actions.
    For each of the groups of actions that should have been merged, based on the above indicators of actions of same type, your thoughts must include the following steps:
    - Find all groups of actions that should have been merged, based on the above indicators of actions of same type.
    - make sure their action type is the same and one of "type", "select", "drag", "click". Otherwise they should not be merged.
    - make sure they are in same app, otherwise they should not be merged.
    - If the action type to merge is "type": is earlier action "element" (in terms of frame id) a prefix of immediate later action "element"?
    - If the action type to merge is "select": is earlier action "element" (in terms of frame id) a subset OR superset of immediate later action "element"?
    - Does this group of actions needs further correction by removing some actions and/or adding more actions? Make sure to go back to the supporting UI change events (find them by ids indicated in "evidences" and "region" fields) of each action under consideration to see if it's to be added or removed from the group.
    - Avoid "over-merging": do not merge actions into one while they actually represent more than one action. Check the supporting UI events of each action in the group to see if they are one or more actions in reality.
    - If the group leads to an "over-merging" the number of true actions is still less than the number of actions in the group, then remember to reason in the same step-by-setp fashion later on each of these subgroups, each corresponding to a true action.
    - write down a conclusion: is this group indeed to merge?

    After your thinking, output the complete and valid JSON object of actions with each of identified group merged, and all other action intact.
    \end{Verbatim}
    \\
    \bottomrule
    \end{tabular}
    \caption{Prompt of Action Corrector : 4/4 (continued)}
    \label{prompt:action-corrector-4}
\end{table*}






\end{appendices}
\end{document}